\documentclass{article}

\PassOptionsToPackage{numbers, compress}{natbib}

\usepackage[preprint]{neurips_2023}

\usepackage[utf8]{inputenc} 
\usepackage[T1]{fontenc}    
\usepackage{hyperref}       
\usepackage{url}            
\usepackage{booktabs}       
\usepackage{amsfonts}       
\usepackage{nicefrac}       
\usepackage{microtype}      
\usepackage{xcolor}         
\usepackage{amsmath}
\usepackage{graphicx}
\usepackage{caption}
\usepackage{subcaption}
\usepackage{xfrac}
\usepackage{wrapfig}

\usepackage{amsthm}
\newtheorem{theorem}{Theorem}

\newtheorem{lemma}{Lemma}

\newtheorem{property}{Property}
\newtheorem*{property*}{Property}

\newtheorem{definition}{Definition}

\newcommand{\E}[2]{\b{E}_{#1}\hspace{-1pt}\left[ #2\right]}
\renewcommand{\b}[1]{\mathbb{#1}}
\renewcommand{\c}[1]{\mathcal{#1}}

\title{Generalization within \textit{in silico} screening}

\author{%
  Andreas Loukas\\
  Roche \\
  \And
  Pan Kessel \\
  Roche \\
  \And
  Vladimir Gligorijevic \\
  Genentech \\
  \And
  Richard Bonneau \\
  Genentech \\
}

\begin{document}
\maketitle

\begin{abstract}
\textit{In silico} screening uses predictive models to select a batch of compounds with favorable properties from a library for experimental validation. Unlike conventional learning paradigms, success in this context is measured by the performance of the predictive model on the selected subset of compounds rather than the entire set of predictions. 
By extending learning theory, we show that the selectivity of the selection policy can significantly impact generalization, with a higher risk of errors occurring when exclusively selecting predicted positives and when targeting rare properties. 
Our analysis suggests a way to mitigate these challenges. 
We show that generalization can be markedly enhanced when considering a model's ability to predict the fraction of desired outcomes in a batch. This is promising, as the primary aim of screening is not necessarily to pinpoint the label of each compound individually, but rather to assemble a batch enriched for desirable compounds. 
Our theoretical insights are empirically validated across diverse tasks, architectures, and screening scenarios, underscoring their applicability.

\end{abstract}

\section{Introduction}

Does a model generalize differently when its own decisions influence on which part of the test set it is evaluated on? Generalization is arguably the most crucial property of any supervised machine learning algorithm. Given this importance, it is no surprise that learning theory has captivated researchers, resulting in a comprehensive body of theoretical results on how predictive models perform on i.i.d. sampled test sets. 

We here shift our focus from the conventional supervised learning perspective to consider the problem of \textit{in silico} (or virtual) screening~\cite{bryant2021deep,gligorijevic2021function,shin2021protein,NEURIPS2023_801ec05b,10.1093/abt/tbab011,stanton2022accelerating,rai2023low,li2023machine,krause2023improving}. Our investigation pertains to scenarios where predictive models are employed to select a batch of compounds, often in anticipation of costly experimental procedures. This batch selection scenario arises prominently in material and drug design, where the goal is to identify compounds with desirable properties, such as low toxicity and target-specific activity.
This identification is typically done via combining predictive models and experiments with fixed batch sizes (with batch size determined by cost and other model-independent factors). Before submitting a batch of designs for experimental validation, we use predictive models to improve the probability that some fraction of the proposed designs has the desired property.
In this context, one employs predictive models to formulate a selection policy determining the likelihood of each compound being chosen for experimental testing. Subsequently, the model's performance is evaluated on the policy-dependent test distribution. 

Through an extension of learning theory arguments, we reveal that the generalization capacity can be notably influenced by the policy's degree of selectivity. 
Particularly, we observe that the a risk of poor generalization arises when exclusively choosing compounds confidently predicted as desirable (highest selectivity). 
Moreover, our analysis illuminates the challenges encountered in discerning `needles in a haystack', where compounds with exceedingly rare \textit{predicted} properties pose formidable obstacles to generalization.

We further broaden our analysis to encompass scenarios where predictive models are used to evaluate the collective quality of a set of examples, exemplified by a classifier's capability to anticipate the fraction of positives within a batch. 
Termed \textit{batched prediction}, this paradigm shifts the focus from individual sample predictions to forecasting the mean label of a batch of examples.  
This approach stems directly from the the primary objective of \textit{in silico} screening, which is not confined to predictions for individual compounds, but emphasizes the assembly of batches enriched with compounds possessing the desired properties. Encouragingly, our findings indicate a marked enhancement in generalization as the batch size increases. This is good news for practitioners who prioritize identifying a high-quality set of designs over demanding guarantees for individual predictions.

Our theoretical insights are examined across different architectures (graph neural networks, convolutional neural networks), application domains (antibody design, quantum chemistry), losses, and tasks (regression of quantum mechanical properties and classification of target binding). Our numerical results highlight that generalization can be affected by the selectivity of the selection policy and that the batched generalization error is reduced significantly with increasing prediction batch size.

\begin{figure}[t]
\centering
\includegraphics[trim={2cm 9cm 2cm 1.85cm},clip,width=0.9\textwidth]{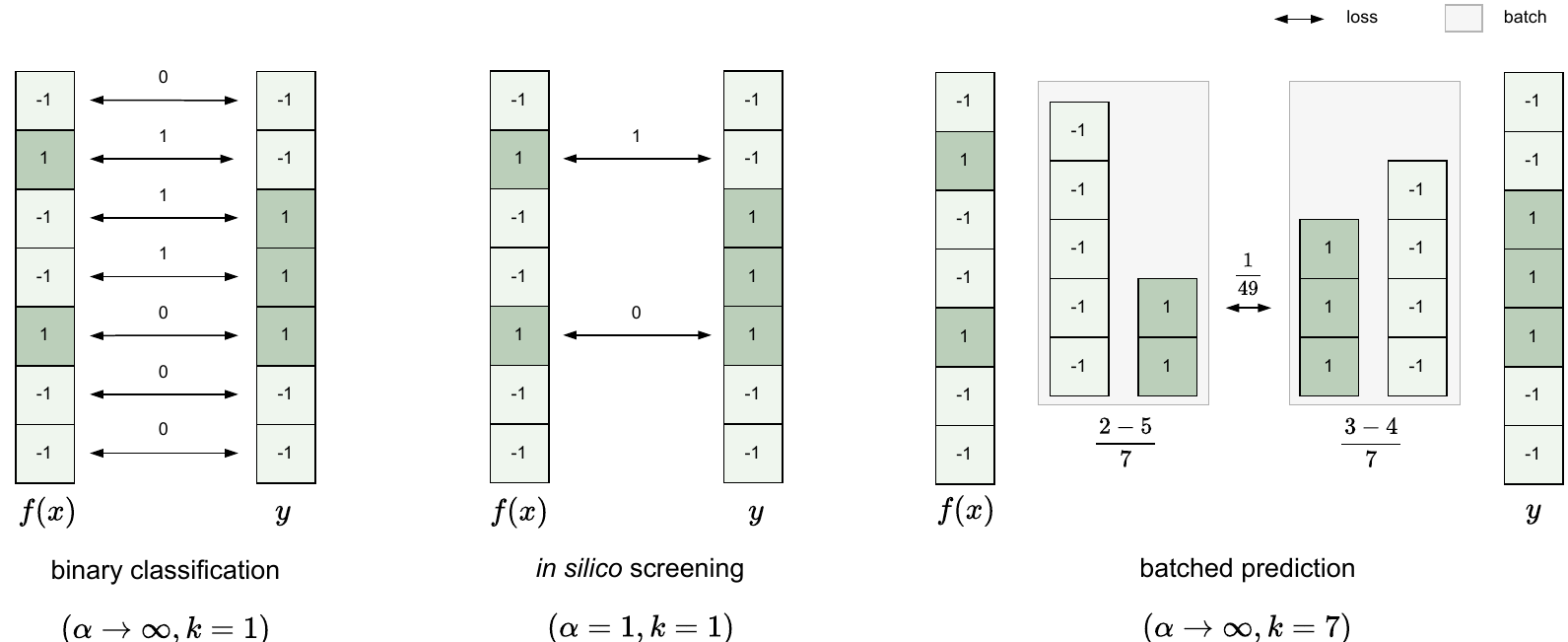}
\caption{Illustration of the different learning paradigms for the case of binary classification with loss $l(f(x),y) = (f(x) - y)^2/4$. The left-most figure depicts the standard learning setting where the prediction $f(x)$ for each example $x$ is compared to the ground truth label $y$. The middle figure exemplifies \textit{in silico} screening with a very selective policy, considering only predicted positives (though we also consider less strict policies in the manuscript). The right-most figure shows batched prediction with a uniform selection policy and a batch size of $k=7$, comparing the total number of positives and negatives in the prediction batch.}
\vspace{-3mm}
\label{fig:illustration}
\end{figure}

\section{Generalization theory for \textit{in silico} screening}

We begin by formally introducing the \textit{in silico} screening task and by deriving upper bounds on the generalization error of screening models. Our analysis brings insights into the unique challenges of screening (as compared to standard binary classification), identifying the degree of selectivity and the rarity of predicted positives as a potential culprit of poor generalization.  


\subsection{Problem definition}

In standard binary classification, we typically assume that the data $z = (x, y) \in \b{Z} = \b{X} \times \{-1, 1\}$ is independent and identically distributed (i.i.d.) with respect to the data density $p$ supported over the domain $\b{Z}$. 
We then aim to learn a model $f:\b{X} \to [-1, 1]$ that minimizes the expected risk   
\begin{align}
    r(f, p) = \b{E}_{z \sim p} \left[ l ( f(x), y ) \right],
\end{align}
where $l$ is an appropriately chosen loss function, such as the mean-squared error loss $l(\hat{y}, y) = (\hat{y} - y)^2$, the zero-one loss $l(\hat{y}, y) = \frac{1}{2} ( 1 - \hat{y} \, y )$, and the cross-entropy loss $l(\hat{y},y) =  - y \log \hat{y} - (1-y) \log (1-\hat{y})$.
In practice, the expected risk is minimized by minimizing the empirical risk over a training set $Z=\left\{ (x_i, y_i) \right\}^n_{i=1}$:
\begin{align}
    r(f, p_Z) 
    = \b{E}_{z \sim p_Z} \left[ l(f(x), y) \right],
\label{eq:per_sample_exp_risk}
\end{align}
where $p_Z$ is the empirical measure 
$p_Z(z) = \frac1n \sum_{i=1}^n \delta(z-z_i)$ and $\delta$ is the Dirac delta function.

In contrast to the standard case, in screening one uses the predictive model to predict which designs will be selected for downstream experimental validation. 
Thus, the standard screening error rate 
\begin{align}
	s(f; p_Z) = \sum_{z \in Z} \pi_f(x) \, l(f(x), y) 
\end{align}
is obtained by re-weighting the loss by a policy $\pi_f$ so that only selected compounds have non-zero selection probability $\pi_f(x)$ and thus contribute to the computed error.
In this work, we focus on policies that are linear w.r.t. the predicted label: 
\begin{align}
	\pi_f(x) \propto \alpha + f(x) \quad \text{with} \quad \alpha \geq 1 
\end{align}
that interpolate between the greedy policy of selecting predicted positives only ($\alpha=1$) and the uniform policy yielding the standard risk ($\alpha \to \infty$).

 %
%

\subsection{Key ingredients of generalization}

The goal of learning theory is to understand the generalization gap between the expected and empirical risk.
Though there are numerous generalization arguments invented thus far, more advanced theories stand out for going beyond the classical paradigm of uniform convergence aiming to capture not only the capacity of the classifier, but also the nature of the data distribution and of the learning algorithm. 

We will derive our results by unifying and re-purposing ingredients from these proven theories. 

\emph{Ingredient 1: Embracing stochasticity through ensembles.} A key ingredient of our analysis borrowed from PAC-Bayesian theories of learning~\cite{mcallester2013pac,mcallester1998some,10.1145/267460.267466,dziugaite2017computing, neyshabur2018pac, zhou2018nonvacuous,NEURIPS2020_ec79d4be,dziugaite2021role,lotfi2022pac} entails considering not a single classifier, but rather a distribution of classifiers determined from the training data. We thus focus on the expected and empirical screening risks of the learned distribution, given respectively by 
\begin{align}
    s(q_Z, p) 
    = \sum_{f \in \c{F}} q_Z(f) s(f; p)
    \quad  \text{and} \quad  
    s(q_Z, p_Z) 
    = \sum_{f \in \c{F}} q_Z(f) s(f; p_Z),
\end{align}
with the expected $r(q_Z, p)$ and empirical risks $r(q_Z, p_Z)$ defined analogously. Above, $\c{F}$ denotes the set of all possible classifiers (usually termed the hypothesis space) and $q_Z$ is a distribution over classifiers learnt by using the training data $Z$. Our choice to consider distributions of classifiers is meant to acknowledge the stochasticity of the training process of neural networks, due to both the random initialization and stochastic optimization. The current setting is intimately connected to the use of classifier ensembles---a common practice in \textit{in silico} screening \cite{tagasovska2023antibody, ani2018virtual, oliveira2023virtual, wu2022prediction}. Crucially, it follows from Jensen's inequality that for any convex loss (w.r.t. the model prediction) the expected risk is an upper bound on the risk of the expected ensemble prediction:
\begin{align}
    s(q_Z; p) 
    \geq \sum_{z} p(x) \, \pi_{q_Z}(x)  \, l\hspace{-1px}\left(\sum_{f \in \c{F}} \frac{q_Z(f) \pi_f(x)}{\pi_{q_Z}(x)} \, f(x), y\right),
\end{align}
with $\pi_{q_Z}(x) = \sum_{f \in \c{F}}q_Z(f)\pi_f(x)$ being the average selection probability as given by the ensemble. Intuitively, the latter corresponds to the error one would obtain if they relied on the average predicted label over the ensemble. Thus, our results will also bound the expected test error of an ensemble. 

\emph{Ingredient 2: Stability of learning.} We further consider the fundamental property of a learning algorithm to be stable to training data perturbations~\cite{bousquet2002stability,hardt2016train, hoffer2017train, kuzborskij2018data, chandramoorthygeneralization}. We will rely on an adaptation of the standard definition that works with distributional ensembles:
\begin{definition}[Stability] 
Let
$
    \beta = n \sup_{Z, Z'} \delta(q_Z, q_{Z'})
$
where the supremum is taken over training sets $Z = \{ z_1, \dots, z_j, \dots, z_n \} \   \textrm{and} \  Z' = \{z_1, \dots, z'_j, \dots, z_n \}$ that differ only at a single example. A learning algorithm that selects a distribution $q_Z$ over classifiers $f \in \c{F}$ is $\beta$-stable if $\beta = o(\sqrt{n})$ grows slower than the square-root of the training set size. 
\end{definition}
Throughout the document, we use the notation $\delta(q,q') = \sum_{x} | q(x) - q'(x)|$ to denote the $\ell_1$ distance between the probability distributions $q$ and $q'$. We note that $\delta(q,q') \leq 2$. 

\emph{Ingredient 3: The Lipschitz constant as a measure of expressive power.} An established way to characterize the expressive power of the set of classifiers that we can learn is by imposing that their output does not change abruptly when one perturbs the input slightly, as quantified by the Lipschitz constant~\citep{von2004distance,xu2012robustness,sokolic2017robust,novak2018sensitivity}. Concretely, a function $f: \c{X} \to [-1, 1]$ has Lipschitz constant $\mu$ within some domain if $|f(x) - f(x')| \leq \mu \|x - x'\|$ for all $x,x' \in \c{X}$. Intuitively, if a classifier has predicted that sample $x$ has label $f(x)=1$ then all samples $x'$ at a distance less than $1/\mu$ from it must also be predicted to be positive as $ f(x') \geq 1 - \mu \, \|x - x'\| >0$. 


\subsection{Selection affects generalization}

Our first technical result upper bounds the expected generalization error $s(q_Z; p)$ of a screening experiment relying on a distribution $q_Z$ of predictive models, implicating the selectivity parameter $\alpha$:
\begin{theorem}
Let $l: [-1, 1] \times [-1, 1] \to [0,1]$ be a Lipschitz continuous loss function with Lipschitz constant $\lambda$. Further, consider a $\beta$-stable learner that uses the training data $Z$ to select a distribution $q_Z$ over hypotheses $f \in \c{F}$ with Lipschitz constant at most $\mu$. The expected screening risk is upper bounded as follows
\begin{align*}
     s(q_Z; p)
    &\leq \frac{ (\alpha + 1) \, \varepsilon_1(\mu, \lambda, p, q, \beta) + \frac{1}{2} \, f_{\text{dev}}(q_Z, p)}{\alpha + f_{\text{avg}}(q_Z, p) }
 \end{align*}
with high probability, where 
\begin{align*}
     \varepsilon_1(\mu, \lambda, p, q, \beta) = \min_q \lambda\mu \, \b{E}_{Z} \left[ \delta(q_Z, q) \, 
 W_1(p, p_Z) \right]  + O\left( \frac{\lambda + \beta}{\sqrt{n}}\right),
\end{align*}
$W_1(p, p_Z)$ measures the 1-Wasserstein distance between the empirical $p_Z$ and true measure $p$ over the labeled data, whereas
\begin{align*}
    f_{\text{avg}}(q_Z, p) = \b{E}_{z\sim p,f \sim q_Z}[f(x)] 
    \quad \text{and} \quad
    f_{\text{dev}}(q_Z, p) = \b{E}_{Z'\sim p^n, f\sim q_Z} \left[\left| \b{E}_{p_{Z'}}[f(x)] - f_{\text{avg}}(q_Z, p) \right|\right]
\end{align*}
are the expected predicted label and the expected absolute deviation of the average predicted positives across different realizations of the training set and ensemble, respectively.  
\label{theorem:generalization_screening_1risk_realizable}
\end{theorem}


A more comprehensive version of Theorem~\ref{theorem:generalization_screening_1risk_realizable} that lifts the assumption that the models fit the training data (referred to as the realizable case) is presented in the supplement. For completeness, we also present variants based on uniform convergence that lift the ensemble and stability assumptions (ingredients 1 and 2) at the expense of less tight bounds.

To comprehend the theorem, we first examine the derived inequality for the standard case of empirical risk that is a special case obtained for $\alpha \to \infty$: 
\begin{align*}
    r(q_Z; p) = \lim_{\alpha\to \infty} s(q_Z; p) 
    \leq \mu \lambda \, \min_q \b{E}_{Z\sim p^n} \left[ \delta\left(q_Z, q\right) \, W_1(p, p_Z) \right] + O\left( \frac{\lambda + \beta}{\sqrt{n}}\right),
\end{align*}
The resulting bound resembles the elegant result of~\citep{NEURIPS2021_4607f7ff} (derived by a Rademacher complexity argument) in its overall form as it captures the dependence on the data distribution by the expected Wasserstein distance between the empirical and true measures and the expressive power of the classifier by the Lipschitz constant $\mu$. However, rather than studying the worst-case behavior of any classifier within the hypothesis class (referred to as the uniform convergence framework), we here follow the PAC-Bayes approach\footnote{The relation to PAC-Bayes bounds becomes more clear by using the inequality $ \delta\left(q_Z, q\right) \leq \sqrt{2 \, KL\left(q_Z, q\right)}$ where the $\ell_1$-distance and KL divergence are upper bounded by 2 and $\infty$, respectively.} of considering the expected behavior under the learned distribution $q_Z$. As a consequence, the obtained bound shrinks when the distance between prior $q$ and posterior $q_Z$ distributions is smaller, matching the intuition that a learning algorithm that adjusts its beliefs less generalizes better. Further, in accordance with previous generalization bounds, the error can be controlled by using a more stable learning algorithm (smaller $\beta$). 

Upon adopting a more selective policy, $\alpha$ with a value near one, the generalization bound changes in two main ways. First, we observe that the frequency of predicted positives $f_{\text{avg}}(q_Z, p)$ plays a large role, with a smaller fraction of predicted positives leading to worse generalization. In particular, the bound becomes vacuous for $\alpha=1$ and $f_{\text{avg}}(q_Z, p)=-1$, i.e., if the classifier only predicts negative labels. We refer to this issue as the `needle in the haystack' problem: we cannot expect the classifier to predict accurately the label of samples that do not appear frequently in its training set. 
Second, we can observe that the generalization bound grows with the ensemble's absolute deviation $f_{\text{dev}}(q_Z, p)$ though the increase is mild. 

\textbf{Practical implications.} 
Our theoretical results therefore suggest that the success of \textit{in silico} screening can be unpredictable unless the library contains a non-negligible fraction of \textit{predicted} positives. Thus, though classifiers can help boost the success rate, we caution against relying on very selective policies to identify compounds with exceedingly rare properties. The silver lining is that we don't need to know the ground truth label distribution to pull back from considering designs from a given library.   


\section{Does generalization improve when evaluating prediction in batches?}

The goal of \textit{in silico} screening is not necessarily to get the prediction of every compound right, but rather to predict how likely it is that a given batch of compounds is going to contain a large fraction with desirable properties. 
Motivated by this observation, Section~\ref{subsec:bached_definition} introduces the batched prediction paradigm, in which the predictor is judged by how well it can predict the average label within a batch. After studying the batched screening risk and the effect of batching on selectivity, we proceed in Section~\ref{subsec:batched_generalization} to analyse how generalization can be improved in this setup.  

\subsection{The batched prediction paradigm}
\label{subsec:bached_definition}

We consider batches  
$S = \left\{ (x_i, y_i) \right\}^k_{i=1}$ of $k$
of data samples drawn i.i.d. from the data distribution with law $p$ and aim to predict the average binary label over the batch 
by the average prediction. 
%
%
We judge the error of a batched predictor based on the expected $k$-risk
\begin{align}
    r_k(f, p) = \b{E}_{S \sim p^k} \left[ l \left( \frac{1}{k} \sum_{z \in S} f(x), \frac{1}{k} \sum_{z \in S} y \right) \right].
\end{align}
where $ \b{E}_{S \sim p^k}$ denotes the expectation over all $k$-batches. 
Similarly, the empirical $k$-risk is defined as
\begin{align}
  r_k(f, p_Z) = \sum_{S \subset_k Z} \frac{1}{\binom{n}{k}} \,  l \left( \frac{1}{k} \sum_{z \in S} f(x), \frac{1}{k} \sum_{z \in S} y \right), \label{eq:standard_batched_empirical}
\end{align}
%
%
where  $S \subset_k Z$ denotes all
%
%
subsets of cardinality $k$ of the training set $Z$ without replacement. Note that $r_k(f,p) = \b{E}_{Z\sim p^n} \left[ r_k(f, p_Z) \right] $.

To appropriately generalize screening to the batched prediction paradigm, we 
suppose that the candidate compounds are drawn from the library in batches and each batch $S$ is selected with probability $\pi_f(S)$ that depends linearly to the number of predicted positives within the batch: 
$$
\pi_f(S) \propto  \alpha + \frac{1}{k} \sum_{z \in S} f(x). 
$$ 
The associated $k$-risk measures the experimental error and is defined as 
\begin{align*}
    s_k(f; p_Z) 
    &= \sum_{S \subset_k Z}  \pi_f(S) \, l\left(\frac{1}{k} \sum_{z \in S} f(x), \frac{1}{k} \sum_{z \in S} y \right)
\end{align*}
and reduces to the standard empirical $k$-risk \eqref{eq:standard_batched_empirical} for the uniform policy $\pi_f(S) = {\binom{n}{k}}^{-1}$.
%

\subsection{Dissecting the effect of batching}
\label{subsec:dissecting_batching}

A property of batching that will turn out to be important for generalization is that risk generally becomes smaller for larger $k$:  
\begin{property}
For the mean-squared error loss (MSE), we have
$$
r_k(f; p_Z) =  r_1(f; p_Z) - \left( 1 - \frac{1}{k}\right) \text{V}_{z \sim p_Z}[e(z)] 
$$
with 
$$
\text{V}_{z \sim p_Z}[e(z)] 
= \frac{n-1}{n^2}\sum_{z \in Z} \left( e(z) - \b{E}_{z \sim p_Z}[e(x)] \right)^2
$$
being the unbiased estimator of the empirical error $e(z) = f(x) - y$ variance.
\label{property:mse_variance}
\end{property}
The above property already hints why we can expect a batched classifier (with a uniform selection policy) to work better: when judging success based on the ability of the classifier to accurately predict the fraction of positives within a batch, the predictor may not be correct on every instance (large variance) and still have low $k$-risk.  More generally, we show that
\begin{property}
For any doubly-convex loss $l$, $k \le k'$, and distribution $p$, 
$
  r_k(f, p) \ge r_{k'}(f, p).
$
\label{property:risk_monotonicity}
\end{property}

On the other hand, to understand the effect of batching on the policy's selectivity, let us suppose that the classifier is confident and predicts integral outputs $\{-1, 1\}$. Then, as derived in Appendix~\ref{app:selectivity}, the expected fraction of predicted positives submitted is 
\begin{align}
    \pi_f(x|f(x)=1)
    &= \frac{\alpha + (1+\alpha) f_{\text{avg}}(Z) + \frac{1}{k}  + \frac{1 - 1/k}{1 - 1/n} \left( f_{\text{avg}}(Z)^2 - 1/n \right)}{2 \, (\alpha + f_{\text{avg}}(Z))}.
\label{eq:selectivity}
\end{align}
The above can be interpreted as the selectivity of the policy. As observed, it decays monotonically with $k$ and $\alpha$ but also depends on the average predicted label $f_{\text{avg}}(Z)$. 

\subsection{Understanding the subtle effects of batching to generalization}
\label{subsec:batched_generalization}

We proceed to bound the generalization error when evaluating prediction in batches. We consider loss functions of the following form:
\begin{align}
  l(\hat{y},y) = K_1(\hat{y},y) \, K_2(\hat{y},y) + b,
\label{def:loss}
\end{align}
where $K_{1}$ and $K_2$ are affine linear functions and $b \in \mathbb{R}$ is a constant shift. We also require that
\begin{align}
 \b{E}_{Z \sim p^n}  \left[   
        l \left( 
        \b{E}_{z \sim p_Z} [f(x)], \b{E}_{z \sim p_Z}[y] 
        \right)  \right]  
        \leq l \left( 
        \b{E}_{z\sim p} [f(x)], \b{E}_{z \sim p}[y] 
        \right) + O\left(\tfrac{1}{n}\right) \, , \label{eq:approx_jensen}
\end{align}
i.e., Jensen's inequality holds for the $n$-risk up to $\sfrac{1}{n}$ corrections.
 Examples of such loss functions include the mean-squared error and the 0-1 loss.
  
The following theorem extends Theorem~\ref{theorem:generalization_screening_1risk_realizable} to the batched screening setting:
\begin{theorem}
Let $l: \b{X} \times [-1, 1] \to [0,1]$ be a $\lambda$-Lipschitz continuous loss of the form~\eqref{def:loss} satisfying the above requirements. Further, consider a $\beta$-stable learner that uses the training data $Z$ to select a distribution $q_Z$ over hypotheses $f \in \c{F}$ with Lipschitz constant at most $\mu$. The expected screening $k$-risk is with high probability upper bounded as follows
\begin{align*}
     s_k(q_Z; p)
    &\leq \frac{ (\alpha + 1) \, \varepsilon_k(\mu, \lambda, p, q, \beta) + \frac{1}{2} \, f_{\text{dev}}(q_Z, p)}{\alpha + f_{\text{avg}}(q_Z, p) }
 \end{align*}
where 
\begin{align*}
     \varepsilon_k(\mu, \lambda, p, q, \beta) = \min_q \lambda\mu \, \b{E}_{Z\sim p^n} \left[ \delta(q_Z, q) \left( (1 - a_{k,n})
 W_1(p, p_Z) + a_{k,n}  W_1(u, u_Z) \right) \right]  + O\left( \frac{\lambda + \beta}{\sqrt{n}}\right),
\end{align*}
$a_{k,n} = (1-1/k)/(1-1/n)$, $W_1(u, u_Z)$ is the 1-Wasserstein distance between the empirical $u_Z$ and true measure $u$ over the unlabeled data ($u_Z(x) = 1/n$ for $(x,y) \in Z$), whereas $W_1(p, p_Z)$ measures the 1-Wasserstein distance between the labeled data. 
\label{theorem:generalization_screening_krisk_realizable}
\end{theorem}
%

Theorem~\ref{theorem:generalization_screening_krisk_realizable} is generalized in the supplement to models that do not exactly fit the training data. 

Contrasting the results of Theorems~\ref{theorem:generalization_screening_1risk_realizable} and~\ref{theorem:generalization_screening_krisk_realizable} we observe two effects of batching: 

\textit{Faster convergence of empirical measure.} First off, for larger batches, the generalization bound becomes a function of the convergence of the empirical measure on the unlabeled data as opposed to the labeled data (that is true for $k=1$). This is beneficial as term $W_1(p, p_Z)$ is generally larger than $W_1(u, u_Z)$ and can become significantly larger for rapidly changing ground-truth labeling functions $y=f^*(x)$ that harbor a high Lipschitz constant (see Property~\ref{property:convergence_labeled_unlabeled} in the supplement)

\textit{Earlier stopping can improve test error.} The second effect of batching boils down to our ability to better control generalization through early stopping. Theorems~\ref{theorem:generalization_screening_1risk_realizable} and~\ref{theorem:generalization_screening_krisk_realizable} indicate that generalization improves when the distance $\delta(q_Z, q)$ between prior and posterior is smaller---a fact also recognized in the classical case ($\alpha \to \infty, k=1$)~\cite{hardt2016train}. A practical way to control $\delta(q_Z, q)$ is by early stopping (or weight decay), in which case the theorem implies that the faster we stop, the better we will generalize. Early stopping is also a hidden culprit behind good out of distribution generalization (par. 1 pg 30~\citep{eastwood2023spuriosity}).  Typically, the early stopping argument (also well established in the literature in the context of stability and stochastic gradient descent) has a significant shortcoming: since there is an inherent trade-off between generalization error and the ability to fit the training data, though one may generalize better by stopping the training earlier, this may not result in a improvement on test error if the empirical risk remains far from zero. Thankfully, this is not necessarily an issue for larger batches as it can be significantly easier to attain a small value of empirical risk as $k$ increases (see Property~\ref{property:mse_variance}
and~\ref{property:risk_monotonicity}). In short, the larger the batch size, the simpler it is to minimize the $k$-risk, the earlier we can stop the training (while achieving an acceptable training error), and thus the better we may generalize.  

\textbf{Practical implications.} Instead of changing $\alpha$ to obtain the desired selectivity, we recommend foregoing per-sample guarantees, setting $\alpha=1$ and relying on batching with a larger $k$ as per~\eqref{eq:selectivity} to achieve the same selectivity. Our results indicate that the batched training error $s_k(q,p_Z)$ will be provide a more reliable estimate of the test error. 









\section{Empirical validation}

We proceed to empirically examine the effect of selectivity and batching to generalization as suggested by our theoretical analysis. We consider both classification and regression as well as different architectures and application domains.
We use a random i.i.d. split and approximate the error by the difference of the empirical screening risk on the test and training set, employing Monte Carlo sampling. 
%
This experiments are repeated for multiple seeds to establish variance estimates.

\begin{figure}[t]
     \centering
     \begin{subfigure}[b]{0.32\textwidth}
         \centering
         \includegraphics[width=\textwidth]{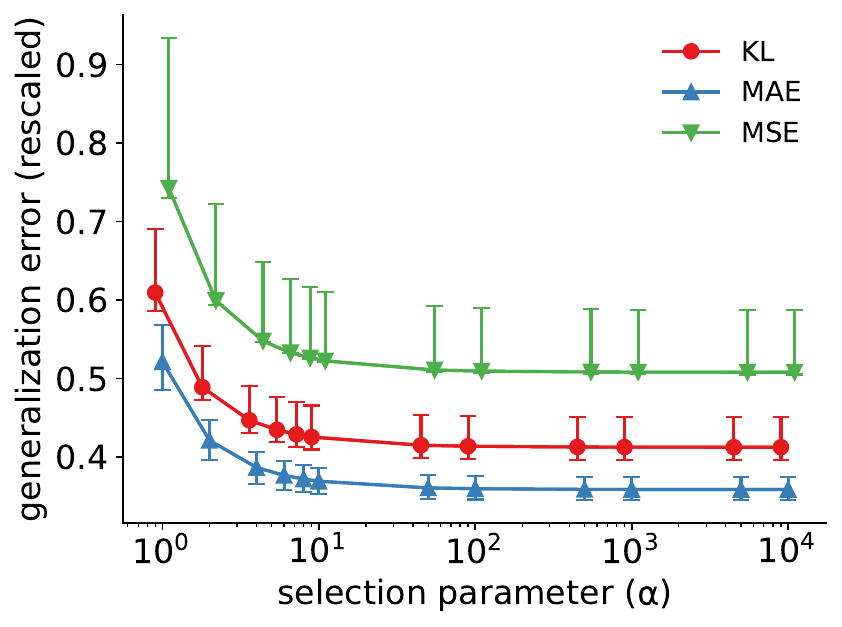}
     \end{subfigure}
     \hfill
     \begin{subfigure}[b]{0.32\textwidth}
         \centering
         \includegraphics[width=\textwidth]{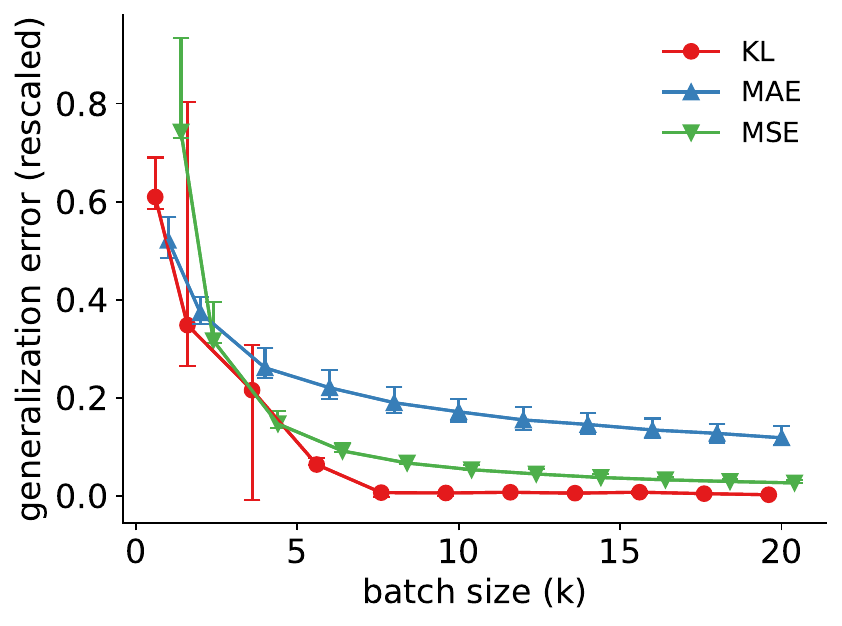}
     \end{subfigure}
     \hfill
     \begin{subfigure}[b]{0.32\textwidth}
         \centering
         \includegraphics[width=\textwidth]{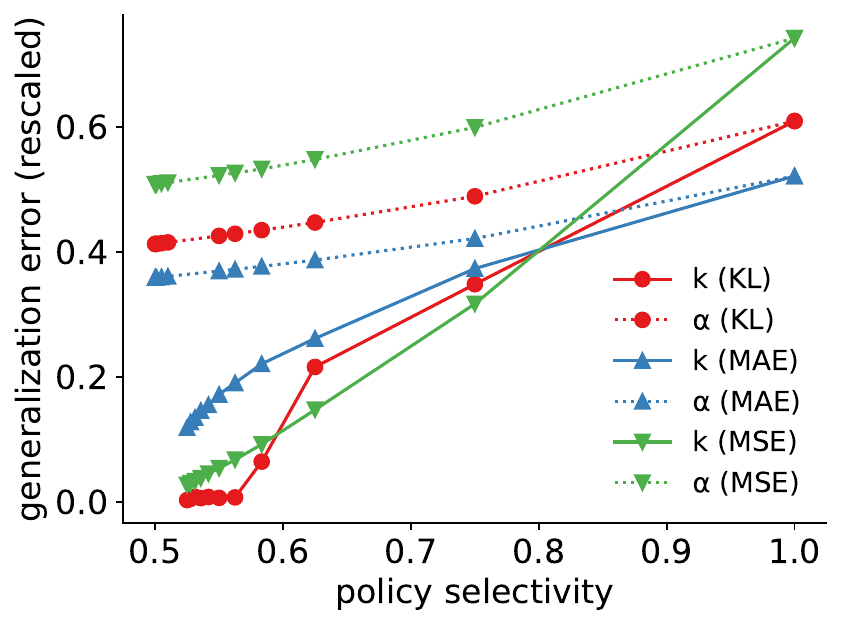}
     \end{subfigure}
      \caption{Estimation of the (rescaled) screening generalization error on the Mason~\textit{et al.} dataset~\citep{mason2021optimization} for different losses as a function of the policy selectivity hyperparameter $\alpha$ (left) and the size of the prediction batch $k$ (middle). Since both $\alpha$ and $k$ influence the policy selectivity, we also compare the generalization error achieved by changing these hyperparameters for the same selectivity. As expected, though both decreasing $\alpha$ and increasing $k$ improves generalization, using a larger batch size yields far superior results.}\label{fig:mason}
\end{figure}

\subsection{Protein design: classification of protein activity} 

We use the Trastuzumab CDR H3 mutant dataset \cite{mason2021optimization, parkinson2023resp} that was constructed by mutating ten positions in the CDR H3 area of the antibody Trastuzumab. The mutants were evaluated in a laboratory to determine binding to the human protein, HER2, a breast cancer target.  The dataset consists of sequences for 9000 binders and 25000 non-binders of which we use the CDR H3 region of the sequence as input to our model. We train a convolutional network to predict if the corresponding antibody binds to the HER2 antigen using a training set of 10k sequences. 
The generalization error is estimated using a test set of 10k sequences. We use binary cross-entropy loss for training. More details on the architecture and training hyperparameters can be found in the Appendix. 

The left-hand side of Figure~\ref{fig:mason} illustrates the rescaled generalization error concerning the KL divergence, the mean average error (MAE), and the mean-squared error (MSE) losses. To ensure comparability, we rescale the loss values by dividing them by the model's test error at initialization ($k=1,\alpha=1$), aligning the three losses on the same scale. Similar to the trend exhibited in our bounds, the screening generalization error diminishes with the increase in both the policy selectivity coefficient $\alpha$ (left) and the prediction batch size $k$ (middle).

As detailed in Section~\ref{subsec:dissecting_batching}, the increase in $k$ also decreases the policy's selectivity. To account for this influence, the rightmost figure contrasts the generalization error while utilizing $\alpha$ and $k$ to regulate selectivity (horizontal axis), as specified in Equation~\eqref{eq:selectivity}. In essence, a selectivity of 1 implies submitting only predicted positives, while a selectivity of 0.5 corresponds to a uniform policy. Remarkably, our observations indicate that batching consistently leads to notably reduced generalization error for the same level of selectivity. This empirical evidence reaffirms our analysis, underscoring that the advantages of batching extend beyond mere selectivity, with significant implications for improving the ability to forecast screening success.



\subsection{Quantum chemistry: regression for molecular property prediction} 

\begin{figure}[t]
    \centering
    \includegraphics[width=1\linewidth]{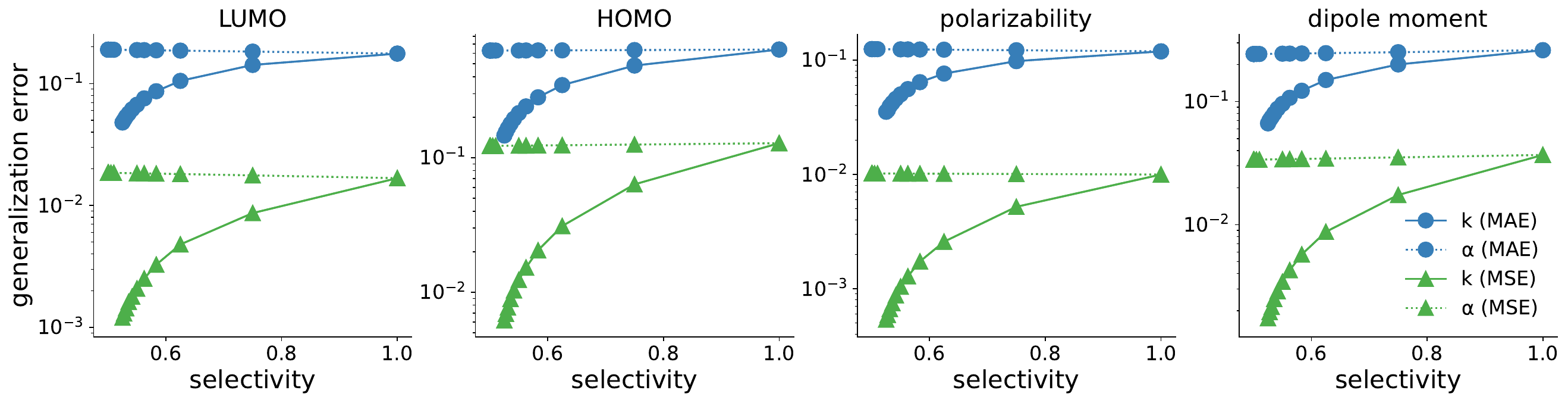}
    \caption{Estimated (normalized) generalization error for QM9 dataset~\cite{wu2017moleculenet} as a function of selectivity. Batched prediction, i.e., varying batch size $k$ at fixed $\alpha$, leads to a lower generalization error than fixing $k=1$ and varying $\alpha$. We train 6 models for each property to estimate the error.}
    \label{fig:qm9_batch_screen}
\end{figure}

We train a graph neural network to predict quantum chemical properties of small molecules. We use the QM9 dataset \cite{wu2017moleculenet} which is one of the standard benchmarks in Quantum Chemistry and consists of 130k molecules along with regression targets, such as the heat capacity and orbital energy of the atomistic system. We use 10k training examples and estimate the generalization error. In order to use a standard graph neural network, we use the same architecture as in the official PyTorch Geometric \cite{pytorchgeom} example for QM9. For a more detailed discussion of the experimental setup, we refer to the Supplementary Material. For consistency and to mitigate the effect of outliers, we rescale all labels to $[-1,1]$ by normalizing based on the 10th and 90th percentiles. Figure~\ref{fig:qm9_batch_screen} shows the normalized estimated generalization error for 4 different properties as a function of the selectivity. We first observe that in this case the gains attained by being less selective are minimal, which is likely due to the label space being balanced after our label normalization. On the other hand, the generalization error decreases by up to an order of magnitude smaller MSE loss, as the batch size $k$ is increased. Hence, we also confirm here that the benefit of batching does not stem from an implicit effect on selectivity, but is an inherent artifact of the $k$-risk.   

\section{Related work}

We next overview how our work relates to existing analyses, applications, and learning paradigms.

\textbf{Study of generalization.} There is an extensive body of work studying the generalization of standard classifiers, e.g., based on VC-dimension~\cite{vapnik1994measuring}, Rademacher complexity~\cite{bartlett2017spectrally}, PAC-Bayes~\cite{mcallester2013pac,mcallester1998some,10.1145/267460.267466,dziugaite2017computing, neyshabur2018pac, zhou2018nonvacuous,NEURIPS2020_ec79d4be,dziugaite2021role,lotfi2022pac}, 
robustness~\cite{xu2012robustness,sokolic2017robust,loukas2021training}, Neural Tangent Kernel \cite{jacot2018neural}, and stability arguments~\cite{bousquet2002stability,hardt2016train, hoffer2017train, kuzborskij2018data, chandramoorthygeneralization}. For a pedagogical review of learning theory, we refer the reader to~\cite{shalev2014understanding}, whereas a modern critical analysis of generalization for deep learning can be found in ~\cite{zhang2021understanding}. To the best of our knowledge, the specific challenges of screening and batched prediction paradigms addressed in this manuscript have not been formally studied before. 

\textbf{Batched screening of molecular libraries.} The tailored design of forecasting models for batched biological sequence classification was previously considered by \cite{wheelock2022forecasting}. Therein, the label posterior of experimental libraries is approximated by a Gaussian mixture model. The practical approach attempts to also account for distribution shift and is numerically shown to improve upon models trained to do standard classification. Our work provides a complementary perspective by elucidating the fundamental generalization limits of standard predictors used for screening.

\textbf{Batching in social sciences.} In the large batch regime $k=O(n)$, batching is being used in social science to draw inferences about populations of documents, such as estimating the prevailing political opinions within
blogs~\citep{HopKin10}. The approach of predicting the population or batch label by averaging individual predictions is well-established. It has also been suggested that the overall prediction quality can be improved further by calibrating the batched predictor using a validation set. The extension of such ideas to provides an avenue for future work.

\textbf{Bayesian optimization (BO).} BO includes methodologies for maximizing an unknown function's value with minimal queries~\citep{frazier2018tutorial, garnett2023bayesian}. Akin to screening, in BO one relies on a training dataset to develop a predictive model for guiding testing selections. However, while our analysis targets the identification of multiple compounds with desired properties in a single experiment, theoretical work in BO seeks to prove that an acquisition strategy will lead to an eventual (or fast) convergence to the optimum (sublinear regret)~\citep{pmlr-v80-scarlett18a, NEURIPS2018_41f860e3, JMLR:v20:18-213, pmlr-v119-nguyen20d, wilson2024stopping}. Further, BO predominantly employs Gaussian processes and assumes regularity in the labeling function. In contrast, we derive bounds dependent on distributions and learning algorithms, consider arbitrary labeling functions and predictive models (including neural networks), and extend our analysis to the batched setting. 



\textbf{Model calibration.} A calibrated classifier provides probabilistic predictions that match the true fraction of positives, quantifying prediction uncertainty~\cite{guo2017calibration, pmlr-v97-song19a, niculescu2005predicting}. In scenarios where $k=n$, batched learning can be seen as a mild form of model calibration, focusing on the mean behavior of the classifier. However, the problems diverge when $k$ is small. Notably, batch prediction has been used to achieve better-calibrated models in the context of tasks like ``odd $k$ out'', where the risk is calculated using batched $k$ cross-entropy loss with averaging in logit space~\cite{muttenthalerset}.




\section{Limitations}
\label{sec:limitations}

Our analysis holds particular relevance in scenarios where predictive models are instrumental in selecting candidates for costly experimental labeling or analysis, such as in the batched biochemical testing of compounds. However, it's essential to acknowledge the limitations of our study: First off, we assume that the test data (before selection) are i.i.d. distributed and do not address the challenge of distribution shift, which frequently arises in practical multi-round applications of these methods (often referred to as lab-in-the-loop setups). In such contexts, each round's selection influences the training data available for subsequent predictive and generative models, inevitably impacting the generalization properties of the predictive model concerning the shifted data distribution. Investigating the implications of this distribution shift offers a promising avenue for future research. In addition, as is the case with most theoretical analyses of generalization, the bounds examined are asymptotic in nature and thus not expected to be tight. Nevertheless, our empirical results validate that the predicted trends featured in the bounds can manifest in practice. 


\textbf{Broader impact.} Though the current work is fundamental in flavor, the topic of \textit{in silico} screening that is studied in this work can have both positive and negative societal impact. One of the primary applications of screening is the development of novel drugs and the identification of materials with enhanced properties, offering substantial benefits to society. Conversely, predictive models utilized in screening can also be employed to select for toxic and pathogenic compounds, presenting risks and ethical considerations. As with any powerful technology, the ramifications of its utilization can vary greatly, contingent upon the intentions and objectives of those deploying it. Thus, careful consideration of the ethical and social implications is imperative as we navigate the application of screening to various domains.

\section{Conclusion}

This paper investigated challenges and opportunities within \textit{in silico} screening, focusing on the interplay between predictive modeling, selection policies, and generalization properties. Through theoretical analyses and empirical validations across diverse scenarios and tasks, our study made several key contributions: we elucidated how the selectivity of selection policies impacts generalization. Additionally, we introduced the concept of batched prediction, demonstrating its efficacy in improving generalization by focusing on the collective quality of compound batches rather than individual predictions.



\bibliography{refs}
\bibliographystyle{unsrt}

\clearpage
\appendix

\section{Details on Numerical Experiments}

\begin{figure}[t!]
  \centering
  \begin{subfigure}[b]{0.32\textwidth}
    \includegraphics[width=\textwidth]{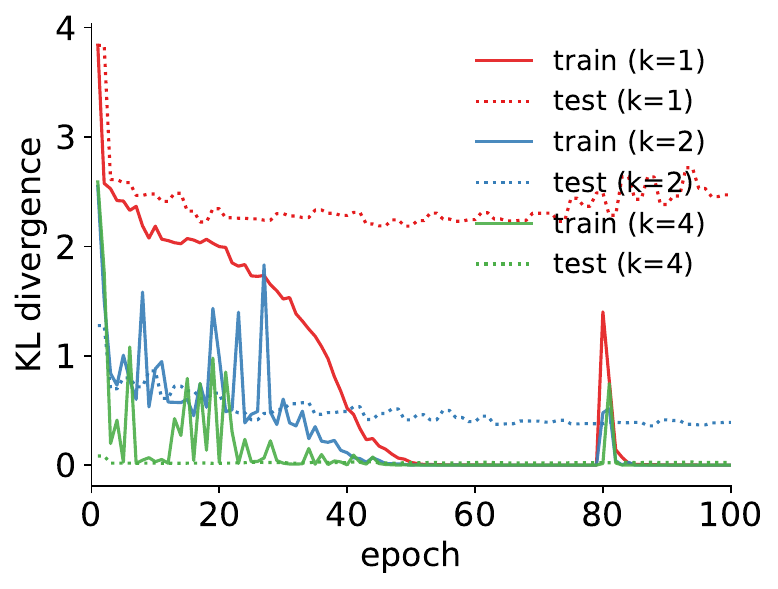}
    \caption{}
    \label{fig:xx}
  \end{subfigure}
  \begin{subfigure}[b]{0.32\textwidth}
    \includegraphics[width=\textwidth]{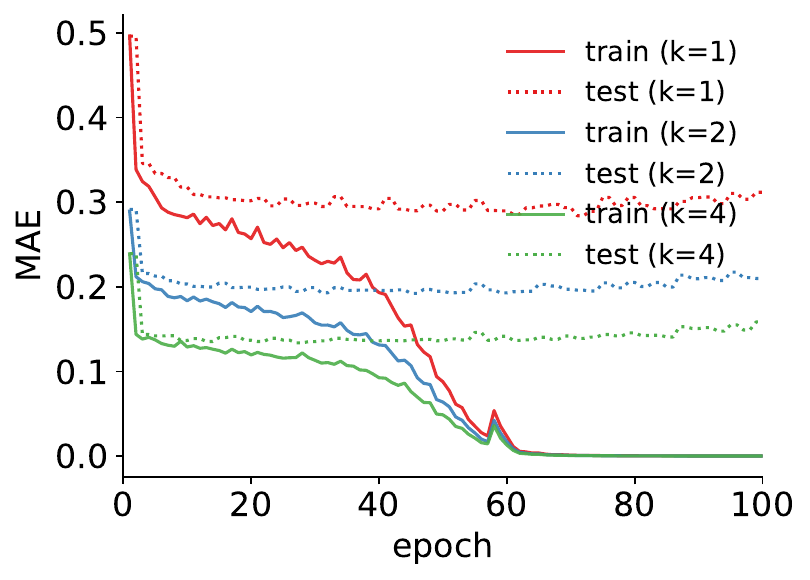}
    \caption{}
    \label{fig:xx}
  \end{subfigure}
  \begin{subfigure}[b]{0.32\textwidth}
    \includegraphics[width=\textwidth]{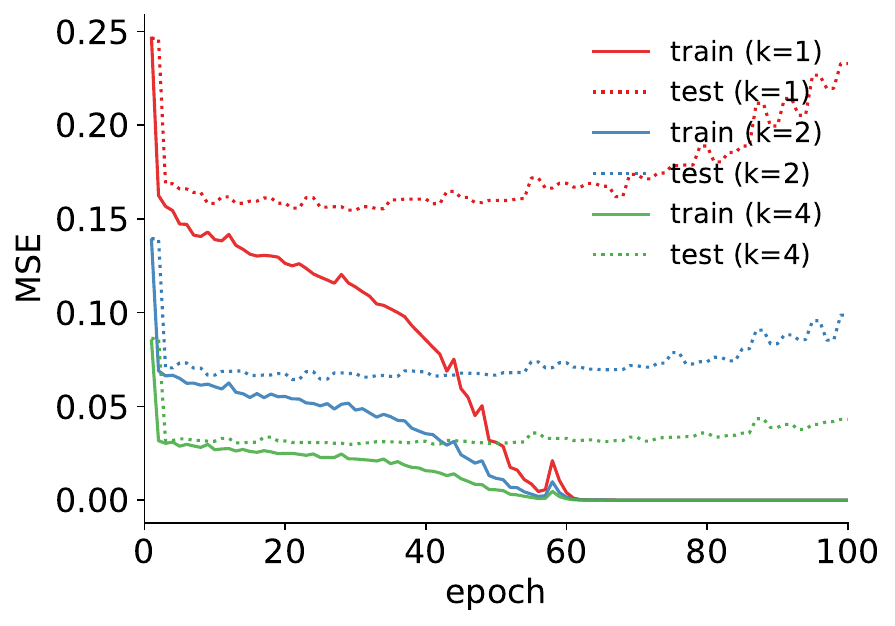}
    \caption{}
    \label{fig:xx}
  \end{subfigure}
  \caption{Training and test losses during training for 3 different batch sizes $k$ in the antibody binding classification task. From left-to-right, the three figures focus on the KL-divergence, MAE and MSE losses, respectively. Both empirical and test risks decrease faster as $k$ increases. The generalization gap reported in the main text is the difference between the train and test loss.}
  \label{fig:mason_training}
\end{figure}

\begin{wrapfigure}{r}{0.36\textwidth}
\centering
    \includegraphics[width=0.35\textwidth]{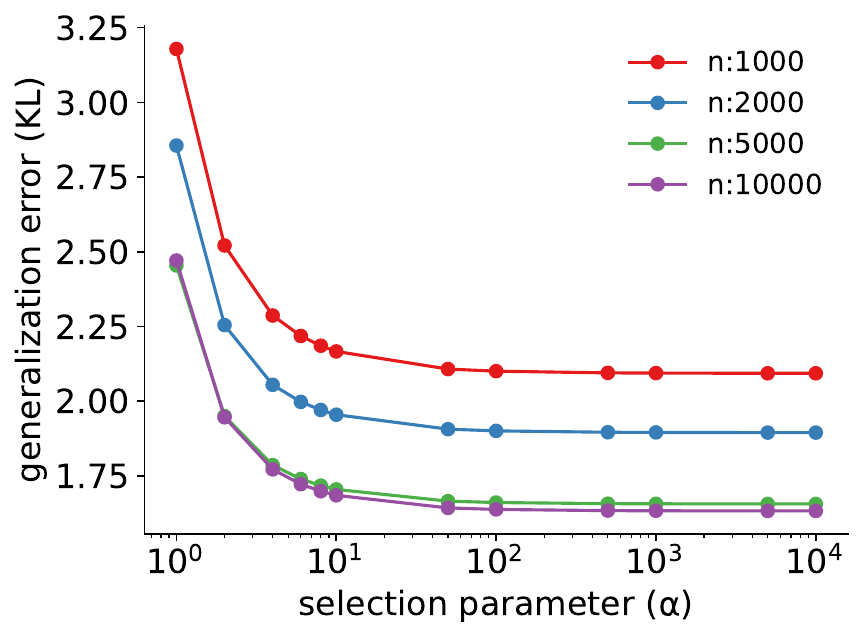}
  \caption{Effect of selectivity parameter $\alpha$ as a function of the training set size $n$. High degree of selectivity remains an issue also for smaller $n$.}
  \label{fig:mason_nsamples}
\end{wrapfigure}
\paragraph{Trastuzumab CDR H3 Mutant Dataset.}
We use the Adam optimizer with a batch-size of 128 with a learning rate of 1e-3 for 100 epochs. We restrict the sequence to the CDR H3 region. We use a 1d-convolutional neural network which uses a 128-dimensional embedding combined with standard positional embedding as input for three one-dimensional convolutions with kernel size 4 and 128 channels. After the last convolutional layer, we take an average over all channels. The output is concatenated with a skip connection of a single linear transformation of the embedded input sequence and then fed into a feed-forward neural network two linear layers of hidden size 128 seperated by a relu activation. For reference, the behavior of the losses during training is depicted in Figure~\ref{fig:mason_training}. Figure~\ref{fig:mason_nsamples} also illustrates the effect of selectivity as a function of the training set size.  

\begin{figure}[t]
\centering
\hspace{14mm}\includegraphics[width=1.0\textwidth]{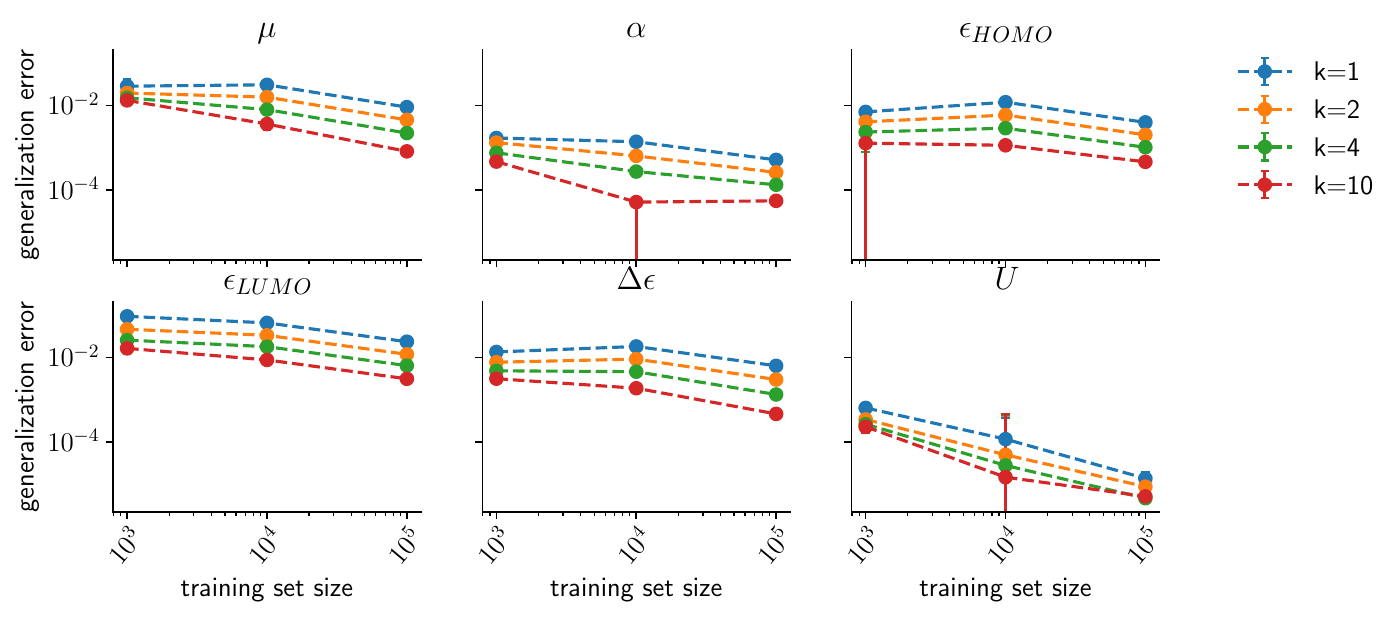}
\caption{Estimated relative generalization error on the QM9 dataset for additional targets. Specifically, we show results for dipole moment $\mu$, isotropic polarizability $\alpha$, highest occupied molecular orbital energy $\epsilon_{HOMO}$, lowest occupied molecular orbital energy $\epsilon_{LUMO}$, HOMO-LUMO gap $\Delta \epsilon = \epsilon_{HOMO} - \epsilon_{LUMO}$, internal energy $U$ at 298.15K.}\label{fig:app_add_qm9}
\end{figure}

\begin{figure}[t]
\centering
\hspace{14mm}\includegraphics[width=1.0\textwidth]{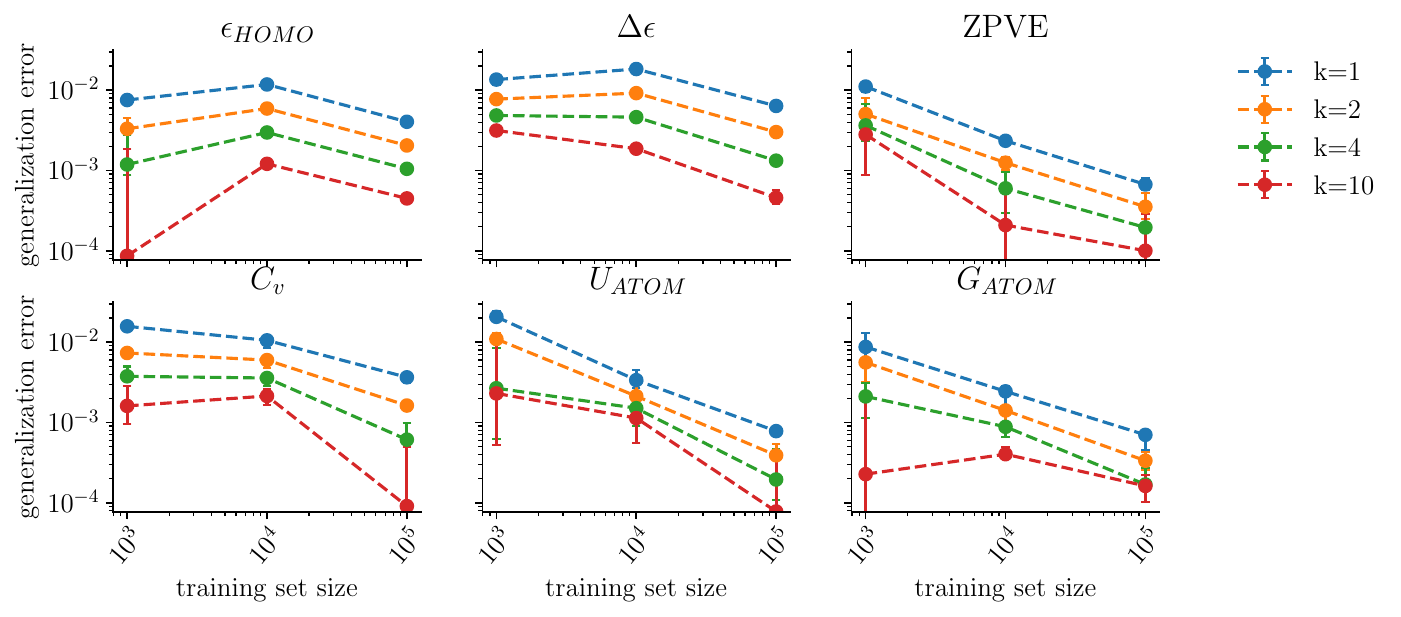}
\caption{Estimated relative generalization error on the QM9 dataset for different training set sizes decreases with the prediction batch sizes $k$ across a wide variety of targets. We show the median of the relative mean absolute error risk, as well as the 25th and 75th percentile over three training runs for each target, initialized with different seeds. The shown targets are the highest occupied molecular orbital energy $\epsilon_{\textrm{HOMO}}$, the gap between highest and lowest occupied orbital energy $\Delta \epsilon$, zero point vibrational energy ZPVE, heat capacity $C_v$, atomization energy $U_{\textrm{ATOM}}$, and atomization enthalpy $G_{\textrm{ATOM}}$. The latter two quantities are with respect to a temperature of 298.15K and the estimate for the last four targets is scaled by a factor of 10 for better visual comparison. }\label{fig:gen_qm9}
\end{figure}

\paragraph{QM9 Dataset.} In order to use a standard architecture, we use the official Pytorch geometric GNN example for QM9 \url{https://github.com/pyg-team/pytorch_geometric/blob/master/examples/qm9_nn_conv.py} with hidden feature size 128. During training, we minimize the mean-squared error loss using the Adam optimizer with learning rate 1e-3 and batch size 128. The estimated generalization error for various targets is shown in Figure~\ref{fig:app_add_qm9} and Figure~\ref{fig:gen_qm9}.

\section{Properties of $k$-risk}

In the following we derive a number of useful properties of $k$-risk that will help us in the analysis of generalization in follow up sections.

\subsection{$k$-risk can be written as a convex combination of 1-risk and $n$-risk}

We consider loss function of form
\begin{align}
  l(h) = K_1(h) K_2 (h) + b , \label{eq:app:loss_form}
\end{align}
with $K_{1}$ and $K_2$ are affine linear in $h\equiv(f(x), y)$ and that $b \in \mathbb{R}$ is a constant shift. We also require that Jensen's inequality holds for the $n$-risk up to $\tfrac1n$ corrections, i.e., 
\begin{align}
 \b{E}_{Z \sim p^n}  \left[   
        l \left( 
        \b{E}_{z \sim p_Z} [f(x)], \b{E}_{z \sim p_Z}[y] 
        \right)  \right]  
        \leq l \left( 
        \b{E}_{z\sim p} [f(x)], \b{E}_{z \sim p}[y] 
        \right) + O\left(\tfrac{1}{n}\right) \, . \label{eq:app_jensen}
\end{align}
We note that many loss functions are of this form stated above. For example, the mean-squared error loss can be obtained by
\begin{align}
b &= 0 , \\
K_1(h) &= K_2(h) =  f(x) - y \,.
\end{align}
Since the mean-squared error loss is convex and thus Jensen's inequality holds exactly, \eqref{eq:app_jensen} holds without any $\tfrac1n$ corrections.
Similarly, the 0-1 loss can be recovered by
\begin{align}
b &= \frac12 , \\
K_1(h) &= -\frac12 f(x) \,, \\ 
K_2(h) &=  y \,.
\end{align}
The approximate Jensen inequality \eqref{eq:app_jensen} for the 0-1 loss can be shown as follows:
\begin{align*}
    \b{E}_{Z \sim p^n} &\left[ l \left( \b{E}_{z\sim p_Z} \left[ f(x) \right], \b{E}_{z\sim p_Z} \left[ y \right] \right) \right] \\ &=
     \frac{1}{2} \left( 1 - \frac{1}{n^2} \, \b{E}_{Z \sim p^n} \left[ \left(\sum_{z \in Z} f(x) \right) \left( \sum_{z \in Z} y \right) \right]  \right) \\
    &= \frac{1}{2} \left( 1 - \frac{n-1}{n} \, \b{E}_{Z \sim p^n} \left[ f(x) \right] \b{E}_{Z \sim p^n} \left[ y \right] - \frac{1}{n} \, \b{E}_{Z \sim p^n} \left[y f(x) \right]  \right)     \\
    &= \frac{1}{2} \left( 1 - \b{E}_{Z \sim p^n} \left[ f(x) \right] \b{E}_{Z \sim p^n} \left[ y \right] \right) + \frac{1}{2n} \left(\b{E}_{Z \sim p^n} \left[ f(x) \right] \b{E}_{Z \sim p^n} \left[ y \right] -  \b{E}_{Z \sim p^n} \left[y f(x) \right]  \right) \\
    &= l \left( \b{E}_{z \sim p} [f(x)], \b{E}_{z \sim p}[y] \right) ] + \frac{1}{2n} \left(\b{E}_{Z\sim p^n} \left[ f(x) \right] \b{E}_{Z\sim p^n} \left[ y \right] -  \b{E}_{Z\sim p^n} \left[y f(x) \right]  \right)
\end{align*}
The last term on the right-hand-side can be bounded by
\begin{align*}
   \frac{1}{2n} \left(\b{E}_{Z\sim p^n} \left[ f(x) \right] \, \b{E}_{Z\sim p^n} \left[ y \right] -  \b{E}_{Z\sim p^n} \left[y  f(x) \right]  \right)
   \leq \frac{1}{n} \,,
\end{align*}
which implies the approximate Jensen inequality \eqref{eq:app_jensen}.

For losses of this type, we can show the following properties:
\begin{property}
  The empirical $k$-risk $r_k(f, p_Z)$ for a loss of the form \eqref{eq:app:loss_form} can be written as convex combination of the empirical $1$-risk and $n$-risk
  \begin{align}
  r_k(f, p_Z) = (1 - a_{k,n}) r_1(f, p_Z) + a_{k, n} \, r_n(f, p_Z)
  \end{align}
  with
  \begin{align}
    a_{k,n} = \frac{n(k-1)}{k(n-1)} \in [0, 1] \,.
  \end{align}
\label{th:convex-comb}
\end{property}
Since $K_{1,2}(h)$ are affine linear with respect to $h$, it holds that
\begin{align}
K_{1,2}\left(\frac{1}{k} \sum_i h_i\right)= \frac1k \sum_i K_{1,2}(h_i) \,. \label{eq:tight-convexity-k}
\end{align}
Note that affine linearity is not only a sufficient but also a necessary condition to fulfill this tight convexity condition.

We first consider the case of a loss function for which $b=0$ for simplicity. We can rewrite the $k$-risk as follows
\begin{align}
r_k(f, p_Z)
&= \hat{\b{E}}_S \left[  K_1\left(\frac1k \sum_{z \in S}h(z)\right) \,  K_2\left(\frac1k \sum_{z' \in S}h(z')\right) \right] \\
&= \hat{\b{E}}_S \left[ \frac1k \sum_{z \in S} K_1(h(z)) \, \frac1k \sum_{z' \in S} K_2(h(z')) \right] \\
&=\hat{\b{E}}_S \left[ \frac1{k^2} \sum_{z \in S} K_1(h(z)) \, K_2(h(z)) \right] + \hat{\b{E}}_S \left[ \frac1{k^2} \sum_{z \neq z' \in S} K_1(h(z)) \, K_2(h(z')) \right] \\
&= \frac1{k} \hat{\b{E}}_z \left[  K_1(h(z)) \, K_2(h(z)) \right] + \frac{k^2-k}{k^2}  \hat{\b{E}}_{z \neq z'} \left[  K_1(h(z)) \, K_2(h(z')) \right] ,
\end{align}
where we have used $\hat{\b{E}}_{S = (z_1, \dots, z_k)} = \hat{\b{E}}_{z_1} \dots \hat{\b{E}}_{z_k}$ in the last step. The last expression above can be rewritten as
\begin{align*}
 &\frac1k r_1(f, p_Z) + \frac{n^2(1-\tfrac1k)}{n(n-1)} \hat{\b{E}}_{z,z'} \left[ K_1(h(z)) \, K_2(h(z')) \right] - \frac{n(1-\tfrac1k)}{n(n-1)} \hat{\b{E}}_z \left[ K_1(h(z)) K_2(h(z))\right] \\
&= \left( \frac1k - \frac{1 - \tfrac1k}{n-1} \right) r_1(f, p_Z) + \frac{n(k-1)}{k(n-1)} \, \frac1n \sum_{z} \, \frac1n \sum_{z'} K_1(z) K_2(z') \,.
\end{align*}
We now use that affine linearity of $K_{1,2}$ implies the tight convexity condition \eqref{eq:tight-convexity-k}. Using this relation, we obtain
\begin{align}
r_k(f, p_Z) &= \left( \frac1k - \frac{1 - \tfrac1k}{n-1} \right) r_1(f, p_Z) + \frac{n(k-1)}{k(n-1)} r_n(f,Z) \\
&= ( 1 - a_{k,n}) r_1(f, p_Z) + a_{k, n} \, r_n(f, p_Z) ,
\end{align}
where we have defined $a_{k,n} =\frac{n(k-1)}{k(n-1)}$.

The above proof for $b=0$ implies immediately for $b\neq0$ that
\begin{align}
r_k(f, p_Z) - b
&= ( 1 - a_{k,n}) (r_1(f, p_Z) - b) + a_{k, n} \, (r_n(f, p_Z) - b) ,
\end{align}
and since $(1 - a_{k,n}) b + a_{k,n} b = b$, it follows that
\begin{align}
r_k(f, p_Z)
&= ( 1 - a_{k,n}) r_1(f, p_Z) + a_{k, n} \, r_n(f, p_Z) ,
\end{align}
for the general case of $b \neq 0$.

\subsection{For doubly-convex loss functions, risk is monotonically decreasing with $k$}

\begin{property*}
  Let $l(x, y)$ be a doubly-convex loss function. Then, it holds that
  \begin{align}
      r_k(f, p) \ge r_{k'}(f, p)
  \end{align}
  for any $k \le k'$.
\end{property*}
\begin{proof}
Note that we can rewrite an average over a set with $k$ elements as an average over all possible sets of $k-1$ elements contained in the set of size $k$:
\begin{align}
    \frac{1}{k} \sum_{i=1}^k f(x_i) = \frac1k \sum_{i=1}^k \frac1{k-1}\sum_{j=1 \atop i \neq j}^k f(x_j) \,.
\end{align}
The expected $k$-risk can be thus written as
\begin{align}
    r_k(f, p) &= \b{E}_{z_1, \dots, z_k} \left[ l \left(\frac1k \sum_{i=1}^k f(x_i), \, \frac{1}{k} \sum_{i=1}^k y_i \right)\right] \\
    &= \b{E}_{z_1, \dots, z_k} \left[ l \left(\frac1k \sum_{i=1}^k \frac1{k-1}\sum_{j=1 \atop i \neq j}^k f(x_j), \, \frac1k \sum_{i=1}^k \frac1{k-1}\sum_{j=1 \atop i \neq j}^k y(x_j) \right)\right]
\end{align}
Since the loss function $l$ is doubly convex
\begin{align}
    l ( \lambda x_1 + (1 - \lambda) x_2, \lambda y_1 + (1 - \lambda) y_2) \le \lambda\, l(x_1, y_1) + (1-\lambda) \, l(x_2, y_2)
\end{align}
for $\lambda \in [0,1]$, we can apply Jensen's inequality to obtain
\begin{align}
    r_k(f, p) &\le \frac1k \sum_{i=1}^k \b{E}_{z_1} \dots \b{E}_{z_k} \left[ \, l \left(\frac1{k-1}\sum_{j=1 \atop i \neq j}^k f(x_j), \frac1{k-1}\sum_{j=1 \atop i \neq j}^k y_j\right) \right] \\
    &= \b{E}_{z_1} \dots \b{E}_{z_{k-1}} \left[ l \left( \frac1{k-1} \sum^{k-1}_{j=1} f(x_j), \frac1{k-1} \sum_{j=1}^{k-1} y_j \right) \right] \\
    &= r_{k-1}(f, p)\,.
\end{align}
We thus obtain
\begin{align}
r_{k}(f, p) \ge r_{k-1}(f, p) \ge \dots \ge r_{k'}(f, p) ,
\end{align}
for any $k \le k'$ as was stated in the theorem.
\end{proof}

\subsection{With an MSE loss, $k$-risk places less priority on the error variance as $k$ increases}

\begin{property*}
For the mean-squared error loss (MSE), we have
$$
r_k(f; p_Z) =  r_1(f; p_Z) - \left( 1 - \frac{1}{k}\right) \text{V}_{z \sim p_Z}[e(z)] 
$$
with 
$$
\text{V}_{z \sim p_Z}[e(z)] 
= \frac{n-1}{n^2}\sum_{z \in Z} \left( e(z) - \b{E}_{z \sim p_Z}[e(x)] \right)^2
$$
being the unbiased estimator of the empirical error $e(z) = f(x) - y$ variance.
\label{lemma:mse_variance}
\end{property*}

We start by expanding the $k$-risk:
\begin{align*}
     r_k(f; p_Z) 
    &= \E{S \subset_k Z}{\left( \sum_{(x,y)\in S} \frac{f(x)}{k} - \sum_{(x,y)\in S} \frac{y}{k} \right)^2 } \tag{$k$-risk} \\
    &= \frac{1}{k^2} \, \E{S \subset_k Z}{\left(\sum_{z\in S} e(z)\right)^2} \\
    &= \E{S \subset_k Z}{ \frac{1}{k^2} \left(\sum_{z\in S} e(z)^2\right)} + \E{S \subset_k Z}{ \frac{1}{k^2} \left(\sum_{z\neq z'\in S} 2 e(z)e(z')\right) }  \\
    &= \frac{1}{k}  r_1(f; p_Z) + \left(1-\frac{1}{k}\right) \E{z\neq z' \sim p_Z}{ e(z)e(z') } \label{eq:app_krist_tmp}
\end{align*}
%
%
The latter term relates to the error variance. To see this write
\begin{align*}
    \frac{n}{n-1} \text{V}_{z \sim p_Z}[e(z)] 
    &= \frac{1}{n}\sum_{z \in Z} \left( e(z) - \b{E}_{z \sim p_Z}[e(z)] \right)^2 \\
    &= \frac{1}{n}\sum_{z \in Z} \left( e(z)^2 +  r_n(f;Z) - 2 \b{E}_{z \sim p_Z}[e(z)] e(z) \right) \\
    &= \frac{1}{n}\sum_{z \in Z}{e(z)^2} +  r_n(f;Z) - 2 \b{E}_{z \sim p_Z}[e(z)] \frac{1}{n}\sum_{z \in Z}{e(z)} \\
    &=  r_1(f; p_Z) -  r_n(f;Z) \\
    &=  r_1(f; p_Z) - \frac{1}{n}  r_1(f; p_Z) - \left(1-\frac{1}{n}\right) \frac{1}{|z\neq z' \in Z|}\sum_{z\neq z' \in Z} { e(z)e(z') } \\
    &= \left(1-\frac{1}{n}\right)  r_1(f; p_Z) - \left(1-\frac{1}{n}\right) \frac{1}{|z\neq z' \in Z|}\sum_{z\neq z' \in Z}{ e(z)e(z') }
\end{align*}
implying
\begin{align*}
     r_1(f; p_Z)
    &= \text{V}_{z \sim p_Z}[e(z)] + \frac{1}{|z\neq z' \in Z|}\sum_{z\neq z' \in Z}{ e(z)e(z') }.
\end{align*}
Substituting the last equation into the expression for the empirical $k$-risk \eqref{eq:app_krist_tmp}, we obtain 
\begin{align*}
     r_k(f; p_Z) 
    &= \frac{1}{k} \text{V}_{z \sim p_Z}[e(z)] +  r_1(f; p_Z) - \text{V}_{z \sim p_Z}[e(z)] \\
    &=  r_1(f; p_Z) - \left( 1 - \frac{1}{k}\right) \text{V}_{z \sim p_Z}[e(z)],
\end{align*}
as claimed.

\section{Generalization for batched predictors}

We recall here the definition of stability

\begin{definition}[stability]
Consider two samples that differ only at a single example: 
\begin{align}
    Z = \{ z_1, \dots, z_j, \dots, z_n \} \quad  \textrm{and} \quad  Z' = \{z_1, \dots, z'_j, \dots, z_n \} .
\end{align}
We say that a learning algorithm that uses the training data $Z$ to select a distribution $q_Z$ over hypotheses $f \in \c{F}$ is $\beta$-stable if
\begin{align}
    \beta := n \sup_{Z, Z'} \delta(q_Z, q_{Z'}) = o(\sqrt{n}),
\end{align}
where $\delta(p,q) = \sum_{x} | p(x) - q(x)|$ is the $\ell_1$ distance between the probability functions. 
\end{definition}


We will prove the following theorem:

\begin{theorem}
Let $l: \b{X} \times [-1, 1] \to [0,1]$ be a $\lambda$-Lipschitz continuous loss fulfilling \eqref{def:loss} and \eqref{eq:app_jensen}. Further, consider a $\beta$-stable learner that uses the training data $Z$ to select a distribution $q_Z$ over hypotheses $f \in \c{F}$ with Lipschitz constant at most $\mu$. The expected $k$-risk is upper bounded as follows
\begin{align*}
     r_k(q_Z, p)
     & \leq r_k(q_Z, p_Z) \\
     &\hspace{-5mm}+ \min_q \frac{\mu}{2} \, \b{E}_{Z \sim p^n} \left[ \delta(q_Z, q) \, \left( (1 - a_{k,n})
 W_1(p, p_Z) + a_{k,n}  W_1(u, u_Z) \right] \right) + O\left( \frac{\lambda + \beta}{\sqrt{n}}\right)
 \end{align*}
with high probability, where $W_1(u, u_Z)$ is the 1-Wasserstein distance between the empirical $u_Z$ and true measure $u$ over the unlabeled data, whereas $W_1(p, p_Z)$ measures the 1-Wasserstein distance between the labeled data. 
\label{theorem:generalization_krisk}
\end{theorem}

\subsection{Bounded differences Lemma}

We first derive the following technical lemmas:

\subsubsection{Worst-case generalization (uniform convergence)}

\begin{lemma}
Let $l$ be a Lipschitz continuous loss function with Lipschitz constant $\lambda$ and $\b{Y} = [-1, 1]$. Then, the maximal generalization error
\begin{align}
    \varphi_k(Z) = \sup_{f \in \c{F}} \left( r_k(f, p) - r_k(f, p_Z) \right) 
\end{align}
obeys the bounded difference condition
\begin{align}
    \left| \varphi_k(Z) - \varphi_k(Z') \right| \le \frac{2 \sqrt{2} \lambda}{n},
\end{align}
where we have used the notation
\begin{align}
    Z = \{ z_1, \dots, z_j, \dots, z_n \} \quad \textrm{and} \quad Z' = \{z_1, \dots, z'_j, \dots, z_n \}.
\end{align}
\label{lemma:bounded_diff_uniform_conv}
\end{lemma}

\begin{proof}
We first rewrite the difference as
\begin{align}
\left|  \varphi_k(Z) - \varphi_k(Z') \right|
&= \left| \sup_{f \in \c{F}} \left( r_k(f, p) - r_k(f, p_Z) \right) - \sup_{f \in \c{F}} \left( r_k(f, p) - r_k(f, p_{Z'}) \right) \right| \\
&\le \sup_{f \in \mathbb{F}} \left| r_k(f,p_Z) - r_k(f,p_{Z'}) \right| ,
\end{align}
where we have used that $\sup_x f_1(x) - \sup_x f_2(x) \le \sup_x (f_1(x) - f_2(x) )$ in the last step. 
The $k$-risk $r_k(f, p_Z)$ involves an average over all possible $k$-subsets of the set $Z$
\begin{align*}
 \left|  \varphi_k(Z) - \varphi_k(Z') \right|
&\le \sup_{f \in \c{F}} \left| \hat{\b{E}}_{S \subset_k Z} \left[ l(\hat{\b{E}}_{z \sim S}\left[ f(x) \right], \hat{\b{E}}_S\left[ y \right] )  \right] - \hat{\b{E}}_{S' \subset_k Z'} \left[ l(\hat{\b{E}}_{S'}\left[ f(x') \right], \hat{\b{E}}_S\left[ y' \right] )  \right] \right|.
\end{align*}
Note that we have introduce the notation $\hat{\b{E}}_X$ to refer to the average w.r.t. elements of a set $X$.
For notational convenience, it is useful to define $g(z) = (f^*(x), y)$, where $f^*$ is the maximizing function of the expression above. We can then rewrite the loss as a function of $g$. The above expression can thus be rewritten more compactly as
\begin{align*}
 \left|  \varphi_k(Z) - \varphi_k(Z') \right|
&\le \left| \hat{\b{E}}_{S \subset_k Z} \left[ \, l(\hat{\b{E}}_S\left[ g(z) \right]) \,  \right] - \hat{\b{E}}_{S' \subset_k Z'} \left[ \, l(\hat{\b{E}}_{S'}\left[ g(z') \right]) \,  \right] \right|.
\end{align*}
In this difference, all $k$-sets which are common between $Z$ and $Z'$ will cancel. We can thus rewrite this expression as an average over sets involving $z_j$ of $Z$ and $z'_j$ of $Z'$. This can be easily done by averaging over all $k-1$-sets of $Z / \{z_j\} = Z' /\{z'_j\}$ as follows
\begin{align*}
 &\left| \varphi_k(Z) - \varphi_k(Z') \right| \\
 & \le \underbrace{\frac{\binom{n-1}{k-1}}{\binom{n}{k}}}_{=\frac{k}{n}} \left| \hat{\b{E}}_{A \subset_{k-1} Z / \{ z_j \} } \left[ l \left(\frac{1}{k} g(z_j) + \frac{k-1}{k} \hat{\b{E}}_A\left[ g(z) \right] \right) - l \left(\frac{1}{k} g(z'_j) + \frac{k-1}{k} \hat{\b{E}}_A\left[ g(z) \right] \right)  \right] \right| , 
\end{align*}
We now use that the loss function $l$ is Lipschitz continuous:
\begin{align}
  \left| l(x + \delta) - l(x + \delta') \right| \le \lambda \,\|\delta - \delta'\|_2 \,.
\end{align}
to obtain
\begin{align*}
 &\left| \varphi_k(Z) - \varphi_k(Z') \right|  \le  \frac{\lambda}{n} \,  \lVert g(z_j) - g(z'_j) \rVert.
\end{align*}
By assumption on boundedness of the hypothesis class and labels, it holds that
\begin{align}
 \lVert g(z_j) - g(z_j') \rVert  = \sqrt{\left(f^*(x_j) - f^*(x'_j)\right)^2 + \left(y_j - y'_j \right)^2 } \le 2 \sqrt{2},
\end{align}
implying also 
\begin{align*}
 &\left| \varphi_k(Z) - \varphi_k(Z') \right|  \le  \frac{\lambda}{n} \, 2 \sqrt{2},
\end{align*}
as desired.
\end{proof}

\subsubsection{Expected generalization of stable learners}

\begin{lemma}
Let $l$ be a Lipschitz continuous loss function with Lipschitz constant $\lambda$ and range $\b{Y} = [-1, 1]$. Further, consider two training sets that differ only at a single example: 
\begin{align}
    Z = \{ z_1, \dots, z_j, \dots, z_n \} \quad  \textrm{and} \quad  Z' = \{z_1, \dots, z'_j, \dots, z_n \}.
\end{align}
The expected generalization error of a $\beta$-stable learning algorithm
\begin{align}
    \psi_k(Z) = \sum_{f \in \c{F}} q_Z(f) \left( r_k(f, p) - r_k(f, p_Z) \right) 
\end{align}
obeys the bounded difference condition
\begin{align}
 &\left| \psi_k(Z) - \psi_k(Z') \right| 
    \le \frac{2 \sqrt{2} \lambda + 2 \beta}{n}. 
\end{align}
\label{lemma:bounded_diff}
\end{lemma}

\begin{proof}
The derivation proceeds by rewriting the difference as
\begin{align}
   \left|  \psi_k(Z) - \psi_k(Z') \right|
   &= \left| \sum_{f \in \c{F}} q_Z(f) \left( r_k(f, p) - r_k(f, p_Z) \right) - \sum_{f \in \c{F}} q_{Z'}(f) \left( r_k(f, p) - r_k(f, p_{Z'}) \right)  \right| \notag \\
   &\hspace{-20mm}\leq \left| \sum_{f \in \c{F}} q_Z(f) \left( r_k(f, p_{Z'}) - r_k(f, p_Z) \right)\right| + \left| \sum_{f \in \c{F}} \left(q_Z(f) - q_{Z'}(f)\right) \left( r_k(f, p) - r_k(f, p_{Z'}) \right)  \right| \notag \\
   &\hspace{-20mm}\le \sup_{f \in \c{F}} \left| r_k(f,p_Z) - r_k(f,p_{Z'}) \right| + \sup_{f \in \c{F}} \left| r_k(f,p) - r_k(f,p_{Z'}) \right| \, \sum_{f \in \c{F}} \left|q_Z(f) - q_{Z'}(f)\right|  \notag \\
   &\hspace{-20mm} \le \frac{2 \sqrt{2} \lambda }{n} + \delta(q_Z, q_{Z'} ) 
   \leq \frac{2 \sqrt{2} \lambda + \beta}{n}, \notag 
\end{align}
where the penultimate step follows from the bounded-ness of the loss and the derivation of  Lemma~\ref{lemma:bounded_diff_uniform_conv}.
\end{proof}

\subsection{Concentration around the mean}

Since $\psi_k$ fulfils the bounded difference condition (special case of Lemma~\ref{lemma:bounded_diff}), we can apply McDiarmid's inequality to obtain
\begin{align}
    \mathbb{P} \left[ \psi_k(Z) - \b{E}_Z \left[ \psi_k(Z) \right] \ge \epsilon \right] \le \exp \left(-\frac{\epsilon^2 n}{c^2}\right)
\end{align}
with $c = 2 \sqrt{2} \lambda + \beta$ (stable learner) or $c = 2 \sqrt{2} \lambda $ (uniform convergence). This probability is below $\delta$ for $\epsilon^* \ge c \sqrt{\frac{\ln(1 / \delta)}{n}}$. This immediately implies that
\begin{align}
    \mathbb{P}\left[ \psi_k(Z) < \b{E}_Z \left[ \psi_k(Z) \right] + \epsilon^* \right] = 1 - \mathbb{P} \left[ \psi_k(Z) - \b{E}_Z \left[ \psi_k(Z) \right]  \ge \epsilon^* \; \right] \ge 1 - \delta,
\end{align}
from which we conclude that the following holds
\begin{align}
    \sum_{f \in \c{F}} q_Z(f) r_k(f, p) 
    \leq \sum_{f \in \c{F}} q_Z(f) \, r_k(f, p_Z) + \b{E}_Z \left[ \psi_k(Z) \right] + c \, \sqrt{\frac{2 \ln (1 / \delta)}{n}} 
\end{align}
with probability at least $1 - \delta$. 


\subsection{The special case of $k=1$}
\label{subsec:generalization_special_case_k=1}

We next derive an upper bound on the generalization error in terms of the Wasserstein distance when the function class contains all $\|f\|_L \leq \mu$ Lipschitz continuous functions.

\subsubsection{Worst-case generalization (uniform convergence)}

In the framework of uniform convergence, we have
\begin{align*}
    \b{E}_{Z \sim p^n} \left[ \varphi_1(Z) \right] 
    &= \b{E}_{Z \sim p^n} \left[\sup_{f \in \c{F}} \left( r_1(f, p) - r_1(f, p_Z) \right) \right] \\
    &= \b{E}_{Z \sim p^n} \left[ \, \sup_{f \in \c{F}} \b{E}_{z \sim p} \left[ l \left( f(x), y \right) \right] - \b{E}_{z \sim p_Z} \left[ l(f(x), y)   \right]  \, \right]
\end{align*}
For any $f \in \c{F}$ and $z$, define $g(z) = l(f(x),y)$ and let $\c{G}$ be the corresponding hypothesis class.
By the sub-multiplicity of Lipschitz constants, we have
$$
    \|g\|_{L} \leq \|l\|_{L} \|f\|_{L} = \lambda \, \|f\|_{L},
$$
meaning also that 
\begin{align*}
    \b{E}_{Z \sim p^n} \left[ \varphi_1(Z) \right] 
    &= \b{E}_{Z \sim p^n} \left[\sup_{\|f\|_{L} \leq \mu} \left( \b{E}_{z \sim p} \left[ l \left( f(x), y \right) \right] - \b{E}_{z \sim p_Z} \left[ l(f(x), y)   \right] \, \right) \, \right] \\
    &\leq \b{E}_{Z \sim p^n} \left[\sup_{\|g\|_{L} \leq \lambda \, \mu} \left( \b{E}_{z \sim p} \left[ g(z) \right] - \b{E}_{z \sim p_Z} \left[ g(z) \right] \, \right) \, \right] \\
    &= \lambda \mu \, \b{E}_{Z \sim p^n} \left[ W_1(p, p_Z) \right] 
\end{align*}
We have thus proved the following generalization bound:
\begin{align}
    r_1(f, p) \le r_1(f, p_Z) + \lambda \mu \, \b{E}_{Z \sim p^n} \left[ W_1(p, p_Z) \right] + 2 \sqrt{2} \lambda \, \sqrt{\frac{2 \ln (1 / \delta)}{n}},
\end{align}
which holds with probability at least $1 - \delta$.

\subsubsection{Expected generalization of stable learners}

%
\begin{align*}
    \b{E}_Z \left[ \psi_1(Z) \right] 
    &= \b{E}_Z \left[\sum_{f \in \c{F}} q_Z(f) \left( r_1(f, p) - r_1(f, p_Z) \right) \right] \\
    &= \min_{q} \b{E}_Z \left[\sum_{f \in \c{F}} \left( q_Z(f) - q(f) + q(f)\right)  \left( r_1(f, p) - r_1(f, p_Z) \right) \right] \\ 
    &= \min_{q} \b{E}_Z \left[\sum_{f \in \c{F}} \left( q_Z(f) - q(f) + q(f)\right)  \left( r_1(f, p) - r_1(f, p_Z) \right) \right] \\     
    &= \min_{q} \sum_{f \in \c{F}} q(f)  \left( r_1(f, p) - \b{E}_Z \left[r_1(f, p_Z)\right] \right)  + \b{E}_Z \left[\sum_{f \in \c{F}} \left( q_Z(f) - q(f)\right)  \left( r_1(f, p) - r_1(f, p_Z) \right) \right] \\     
    &= \min_{q} \b{E}_Z \left[\sum_{f \in \c{F}} \left( q_Z(f) - q(f)\right)  \left( r_1(f, p) - r_1(f, p_Z) \right) \right] \\     
    &\leq \min_{q} \b{E}_Z \left[ \sum_{f \in \c{F}} \left| q_Z(f) - q(f) \right|  \sup_{f \in \c{F}} \left| \b{E}_{z \sim p} \left[ l \left( f(x), y \right) \right] - \b{E}_{z \sim p_Z} \left[ l(f(x), y) \right] \, \right| \, \right] \\
    &= \min_{q} \b{E}_Z \left[ \delta\left(q_Z, q\right) \, \sup_{f \in \c{F}} \left| \b{E}_{z \sim p} \left[ l \left( f(x), y \right) \right] - \b{E}_{z \sim p_Z} \left[ l(f(x), y) \right] \, \right| \, \right]
\end{align*}
For any $f \in \c{F}$ and $z$, define $g(z) = l(f(x),y)$ and let $\c{G}$ be the corresponding hypothesis class.
By the sub-multiplicity of Lipschitz constants, we have
$$
    \|g\|_{L} \leq \|l\|_{L} \|f\|_{L} \leq \mu \lambda,
$$
meaning also that 
\begin{align*}
    \b{E}_Z \left[ \psi_1(Z) \right] 
    &= \b{E}_Z \left[\delta\left(q_Z, q\right) \,  \sup_{\|f\|_{L} \leq \lambda} \left| \b{E}_{z \sim p} \left[ l \left( f(x), y \right) \right] - \b{E}_{z \sim p_Z} \left[ l(f(x), y)   \right] \, \right| \, \right] \\
    &\leq \b{E}_Z \left[\delta\left(q_Z, q\right) \,  \sup_{\|g\|_{L} \leq \mu \lambda} \left| \b{E}_{z \sim p} \left[ g(z) \right] - \b{E}_{z \sim p_Z} \left[ g(z) \right] \, \right| \, \right] \\
\end{align*}
W.l.o.g., we will suppose that our hypothesis class abides to the following property:
$$
    \forall g \in \c{G}, \forall x \in X: \quad \exists g' \in \c{G} \ \text{ with } \ g(x) = -g'(x) 
$$
If the property doesn't hold, then we can simply expand the function class appropriately and then continue the proof with the expanded function class. 

Then, for every $g \in \c{G}$ the following holds:
\begin{align*}
    \left| \frac{1}{n} \sum_{i=1}^n g(z_i) - \b{E}_{z \sim p} [g(z)] \right| 
    &= \max \left\{ \frac{1}{n} \sum_{i=1}^n g(z_i) - \b{E}_{z \sim p} [g(z)], \frac{1}{n} \sum_{i=1}^n (-g(z_i)) - \b{E}_{z \sim p} [(-g(z))]\right\} \\
    &\leq \sup_{g \in \c{G}} \frac{1}{n} \sum_{i=1}^n g(z_i) - \b{E}_{z \sim p} [g(z)],
\end{align*}
meaning that we can get rid of the absolute value in our bound
\begin{align*}
    \b{E}_Z \left[ \psi_1(Z) \right]   &\leq \b{E}_Z \left[\delta\left(q_Z, q\right) \,  \sup_{\|g\|_{L} \leq \mu \lambda} \left( \b{E}_{z \sim p} \left[ g(z) \right] - \b{E}_{z \sim p_Z} \left[ g(z) \right] \, \right) \, \right] \\
    &\leq \mu \lambda \, \b{E}_Z \left[ \delta\left(q_Z, q\right) \, W_1(p, p_Z) \right] 
\end{align*}
We have thus proved the following generalization bound:
\begin{align}
    \b{E}_{f \sim q_Z} \left[r_1(f, p)\right] \leq \b{E}_{f \sim q_Z} \left[ r_1(f, p_Z)\right] + \min_q \mu \lambda \, \b{E}_Z \left[ \delta\left(q_Z, q\right) \, 
 W_1(p, p_Z) \right] + (2 \sqrt{2} \lambda + \beta) \, \sqrt{\frac{2 \ln (1 / \delta)}{n}}, \notag
\end{align}
which holds with probability at least $1 - \delta$, which also implies that the following holds 
\begin{align}
    \b{E}_{f \sim q_Z} \left[r_1(f, p)\right] \leq \b{E}_{f \sim q_Z} \left[ r_1(f, p_Z)\right] + \min_q \mu \lambda \, \b{E}_Z \left[ \delta\left(q_Z, q\right) \, 
 W_1(p, p_Z) \right] + O\left( \beta \sqrt{\frac{\log(1/\delta)}{n}}\right) \notag
\end{align}
with the same probability (over $Z$) for any Lipschitz continuous loss function with $\lambda = O(1)$, hypothesis class with maximum Lipschitz constant $\mu$, and and $\beta$-stable learner.


\subsection{The special case of $k=n$}

\subsubsection{Worst-case generalization (uniform convergence)}

We next focus on the expected generalization error term: 
\begin{align*}
    \b{E}_{Z\sim p^n} \left[ \varphi_n(Z) \right] 
    &= \b{E}_{Z\sim p^n} \left[\sup_{f \in \c{F}} \left( r_n(f, p) - r_n(f, p_Z) \right) \right] \\
    &= \b{E}_{Z\sim p^n} \left[ \, \sup_{f \in \c{F}} \left(  \b{E}_{\tilde{Z} \sim p^n} \left[ l \left( \b{E}_{z \sim p_{\tilde{Z}}} \left[ f(x) \right], \b{E}_{z \sim p_{\tilde{Z}}} \left[ y \right] \right) \right] - l(\b{E}_{z \sim p_Z} \left[ f(x) \right], \b{E}_{z \sim p_Z} \left[ y \right] ) \, \right) \, \right], 
\end{align*}
By the approximate Jensen's inequality \eqref{eq:approx_jensen}, we have 
\begin{align*}
     \b{E}_{\tilde{Z} \sim p^n} \left[ l \left( \b{E}_{z \sim p_{\tilde{Z}}} \left[ f(x) \right], \b{E}_{z \sim p_{\tilde{Z}}} \left[ y \right] \right) \right] 
     \leq l \left( \b{E}_{z \sim p} \left[ f(x) \right], \b{E}_{z \sim p} \left[ y \right] \right) + O(\tfrac{1}{n}).
\end{align*}
%
%
Therefore, we can express the expected maximal generalization error as
\begin{align*}
    \b{E}_{Z \sim p^n} \left[ \varphi_n(Z) \right] 
    &\leq
    \b{E}_{Z \sim p^n} \left[ \sup_{f \in \c{F}} \left(  l \left( \b{E}_{z \sim p} [f(x)], \b{E}_{z \sim p}[y] \right) ] - l \left( \frac{1}{n} \sum_{z \in Z} f(x), \frac{1}{n} \sum_{z \in Z} y \right) \, \right) \, \right] + O(\tfrac{1}{n}).
\end{align*}
%
%
By the Lischitz continuity of the loss and the fact that the $\|x\|_2 \leq \|x\|_1$, we have 
$$
    l(a,c) - l(b,d) \leq \lambda \, (|a - b|^2 + |c-d|^2)^{1/2} \leq \lambda \, (|a - b| + |c-d|)
$$
implying also
\begin{align*}
    \b{E}_{Z\sim p^n} \left[ \varphi_n(Z) \right] 
    &\le \lambda \, \b{E}_{Z \sim p^n} \left[ \, \sup_{f \in \c{F}} \left( \left| \frac{1}{n} \sum_{z \in Z} f(x) - \b{E}_{z \sim p} [f(x)]\right| + \left|\b{E}_{z \sim p} [y]- \frac{1}{n} \sum_{z \in Z} y \right| \right) \, \right] + O(\tfrac{1}{n})\\
    &\le \lambda \, \b{E}_{Z \sim p^n} \left[ \, \sup_{f \in \c{F}} \left| \frac{1}{n} \sum_{z \in Z} f(x) - \b{E}_{z \sim p} [f(x)]\right| \, \right] + \lambda \, \left|\b{E}_{z \sim p} [y] - \frac{1}{n} \sum_{z \in Z} y \right| + O(\tfrac{1}{n}).
\end{align*}
Let us focus on the left-most term. 
%


W.l.o.g., we will suppose that our hypothesis class abides to the following property:
$$
    \forall f \in \c{F}, \forall x \in X: \quad \exists f' \in \c{F} \ \text{ with } \ f(x) = -f'(x) 
$$
If the property doesn't hold, then we can simply expand the function class appropriately and then continue the proof with the expanded function class. 

Then, for every $f \in \c{F}$ the following holds:
\begin{align*}
    \left| \frac{1}{n} \sum_{i=1}^n f(x_i) - \b{E}_{z \sim p} [f(x)] \right| 
    &= \max \left\{ \frac{1}{n} \sum_{i=1}^n f(x_i) - \b{E}_{z \sim p} [f(x)], \frac{1}{n} \sum_{i=1}^n (-f(x_i)) - \b{E}_{z \sim p} [(-f(x))]\right\} \\
    &\leq \sup_{f \in \c{F}} \frac{1}{n} \sum_{i=1}^n f(x_i) - \b{E}_{z \sim p} [f(x)],
\end{align*}
meaning that we can get rid of the absolute value in our bound:
\begin{align*}
    A := \b{E}_{Z} \left[ \, \sup_{f \in \c{F}} \left| \frac{1}{n} \sum_{i=1}^n f(x_i) - \b{E}_{z \sim p} [f(x)]\right| \, \right] 
    &= \b{E}_{Z} \left[ \, \sup_{f \in \c{F}} \left( \frac{1}{n} \sum_{i=1}^n f(x_i) - \b{E}_{z \sim p} [f(x)] \right) \right]
\end{align*}

We recall the Kantorovich-Rubenstein duality theorem, which states that when the probability space is a metric space, then for any fixed $\mu > 0$,
\begin{align*}
    W_1(p, p') = \frac{1}{\mu} \sup_{\|f\|_L \leq \mu} \b{E}_{x \sim p} [f(x)] - \b{E}_{x \sim p'} [f(x)],
\end{align*}
with $\|f\|_L$ being the Lipschitz constant of $f$.

Thus, imposing a $\mu$ Lipschitz-continuity assumption on the function class gives 
\begin{align*}
    A 
    &= \b{E}_{Z \sim p^n} \left[ \, \sup_{\|f\|_L \leq \mu} \b{E}_{x \sim u_Z} [f(x)] - \b{E}_{x \sim u} [f(x)] \right] 
    = \mu \, \b{E}_{Z \sim p^n} \left[ W_1(u_Z, u) \right],
\end{align*} 
which is the expected 1-Wasserstein distance between two empirical measures of $n$ samples multiplied by the Lipschitz constant $\mu$.
%

The following Lemma controls the difference of labels term $B$: 
\begin{lemma}
For $y_1, \ldots, y_n$ i.i.d. Rademacher random variables: $\b{E} \left[ \left| \frac{1}{n} \sum_{i=1}^n y_i - \b{E}[y] \right| \, \right] \leq \sqrt{\frac{1}{n}}$.
\label{lemma:difference_of_labels}
\end{lemma}

\begin{proof}
We invoke Jensen's inequality to obtain an upper bound in terms of the variance of a random variable.
\begin{align*}
    \left(\b{E} \left[ \left| \frac{1}{n} \sum_{i=1}^n y_i - \b{E}[y] \right| \, \right]\right)^2 
    &\leq \b{E}\left[ \left( \frac{1}{n} \sum_{i=1}^n y_i - \b{E}[y] \right)^2 \, \right] \\
    &= \mathbb{V}(Y) \tag{set $Y = \frac{1}{n} \sum_{i=1}^n y_i$}
\end{align*}
The random variable 
$$
    n (Y+1)/2 = \sum_{i=1}^n (y_i + 1)/2
$$
has a binomial law and thus $\mathbb{V}(n (Y+1)/2) = n p (1-p) \leq n/4$. It follows that
\begin{align*}
    \b{E} \left[ \left| \frac{1}{n} \sum_{i=1}^n y_i - \b{E}[y] \right| \, \right]
    &\leq \sqrt{\mathbb{V}(Y)} 
    = \sqrt{\mathbb{V}(Y+1)} 
    = \sqrt{\frac{4}{n^2} \mathbb{V}(n (Y+1)/2)} 
    \leq \sqrt{\frac{1}{n}} 
\end{align*}
\end{proof}

Combining the bounds on $A$ and $B$ results to 
\begin{align*}
    \b{E}_Z \left[ \varphi_n(Z) \right] 
    &\leq \lambda \mu \, \b{E}_{Z \sim p^n} \left[ W_1(u_Z, u) \right] + \lambda 
 \sqrt{\frac{1}{n}} + O(\tfrac{1}{n})
\end{align*}
and thus with at least probability $1 - \delta$:
\begin{align}
    r_n(f, p) 
    &\le r_n(f, p_Z) + \lambda \mu \, \b{E}_{Z \sim p^n} \left[ W_1(u_Z, u) \right] + \lambda \sqrt{\frac{1}{n}} + 2 \sqrt{2} \lambda \sqrt{\frac{2 \log (1 / \delta)}{n}} + O(\tfrac{1}{n}) \notag
\end{align}
%
%
The latter implies that the following holds 
\begin{align}
    r_n(f, p) 
&\le r_n(f, p_Z) + \lambda \mu \, \b{E}_{Z \sim p^n} \left[ W_1(u_Z, u) \right] + O\left(\sqrt{\frac{\log(1/\delta)}{n}}\right)\notag
\end{align}
holds with the same probability assuming $\lambda = O(1)$.

\subsubsection{Expected generalization of stable learners}


We next focus on the expected generalization error term: 
\begin{align*}
    \b{E}_{Z \sim p^n} \left[ \psi_n(Z) \right] 
    &= \b{E}_{Z\sim p^n} \left[\sum_{f \in \c{F}} q_Z(f) \left( r_n(f, p) - r_n(f, p_Z) \right) \right] \\
    &= \b{E}_{Z\sim p^n} \left[ \sum_{f \in \c{F}} q_Z(f) \left(  \b{E}_{\tilde{Z}\sim p^n} \left[ l \left( \b{E}_{z \sim p_{\tilde{Z}}} \left[ f(x) \right], \b{E}_{z \sim p_{\tilde{Z}}} \left[ y \right] \right) \right] - l(\b{E}_{z \sim p_Z} \left[ f(x) \right], \b{E}_{z \sim p_Z} \left[ y \right] ) \, \right) \, \right] \\
    &\leq \b{E}_{Z\sim p^n} \left[ \sum_{f \in \c{F}} q_Z(f) \left(  l \left( \b{E}_{z \sim p} \left[ f(x) \right], \b{E}_{z \sim p} \left[ y \right] \right) - l(\b{E}_{z \sim p_Z} \left[ f(x) \right], \b{E}_{z \sim p_Z} \left[ y \right] ) \, \right) \, \right] + O(\tfrac{1}{n}) \,.
\end{align*}
where, once more, we use the approximate Jensen's inequality \eqref{eq:approx_jensen} to pass the expectation within the loss.
Assuming further that our loss abides to 
$
    l(a,c) - l(b,d) \leq \lambda \,  (|a - b| + |c-d|)
$
which (as shown above) is a consequence of Lipschitz continuity, yields
\begin{align*}
    \b{E}_{Z\sim p^n} \left[ \psi_n(Z) \right] 
    &\le \lambda \, \b{E}_{Z \sim p^n} \left[ \sum_{f \in \c{F}} q_Z(f) \left( \left| \frac{1}{n} \sum_{z \in Z} f(x) - \b{E}_{z\sim p} [f(x)]\right| + \left|\b{E}_{z\sim p} [y]- \frac{1}{n} \sum_{z \in Z} y \right| \right) \, \right] + O(\tfrac{1}{n}) \\
    &= \lambda \, \b{E}_{Z \sim p^n} \left[\sum_{f \in \c{F}} q_Z(f) \left| \frac{1}{n} \sum_{z \in Z} f(x) - \b{E}_{z\sim p} [f(x)]\right| \, \right] + \lambda \, \left|\b{E}_{z\sim p} [y] - \frac{1}{n} \sum_{z \in Z} y \right| + O(\tfrac{1}{n}) \,. \\
\end{align*}
Let us focus on the left-most term:
\begin{align*}
    A &:= 
    \b{E}_{Z} \left[\sum_{f \in \c{F}} q_Z(f) \left| \frac{1}{n} \sum_{z \in Z} f(x) - \b{E}_{z\sim p} [f(x)]\right| \, \right]  \\
    &=\min_q \b{E}_{Z} \left[\sum_{f \in \c{F}} (q_Z(f)-q(f)) \left| \frac{1}{n} \sum_{z \in Z} f(x) - \b{E}_{z\sim p} [f(x)]\right| \, \right] + \b{E}_{Z} \left[\sum_{f \in \c{F}} q(f) \left| \frac{1}{n} \sum_{z \in Z} f(x) - \b{E}_{z\sim p} [f(x)]\right| \, \right] \\
    &=\min_q \b{E}_{Z} \left[\sum_{f \in \c{F}} (q_Z(f)-q(f)) \left| \frac{1}{n} \sum_{z \in Z} f(x) - \b{E}_{z\sim p} [f(x)]\right| \, \right] + \sum_{f \in \c{F}} q(f) \b{E}_{Z} \left[\left| \frac{1}{n} \sum_{z \in Z} f(x) - \b{E}_{z\sim p} [f(x)]\right| \right] \\
    &=\min_q \b{E}_{Z} \left[\sum_{f \in \c{F}} (q_Z(f)-q(f)) \left| \frac{1}{n} \sum_{z \in Z} f(x) - \b{E}_{z\sim p} [f(x)]\right| \, \right] + \sqrt{\frac{1}{n}} \tag{from Lemma~\ref{lemma:difference_of_labels}} \\
    &\leq \min_q \b{E}_{Z} \left[ \delta(q_Z, q) \, \sup_f \left| \frac{1}{n} \sum_{z \in Z} f(x) - \b{E}_{z\sim p} [f(x)]\right| \, \right] + \sqrt{\frac{1}{n}} \,.
\end{align*}


Without loss of generality, we assume that our hypothesis class abides to the following property:
$$
    \forall f \in \c{F}, \forall x \in X: \quad \exists f' \in \c{F} \ \text{ with } \ f(x) = -f(x) 
$$
If the property doesn't hold, then we can simply expand the function class appropriately and then continue the proof with the expanded function class. 

Then, for every $f \in \c{F}$ the following holds:
\begin{align*}
    \left| \frac{1}{n} \sum_{i=1}^n f(x_i) - \b{E}_{z\sim p} [f(x)] \right| 
    &= \max \left\{ \frac{1}{n} \sum_{i=1}^n f(x_i) - \b{E}_{z\sim p} [f(x)], \frac{1}{n} \sum_{i=1}^n (-f(x_i)) - \b{E}_{z\sim p} [(-f(x))]\right\} \\
    &\leq \sup_{f \in \c{F}} \frac{1}{n} \sum_{i=1}^n f(x_i) - \b{E}_{z\sim p} [f(x)],
\end{align*}
meaning that we can get rid of the absolute value in our bound:
\begin{align*}
    A \leq \min_q \b{E}_{Z} \left[  \delta(q_Z, q) \, \sup_{f \in \c{F}} \left( \frac{1}{n} \sum_{i=1}^n f(x_i) - \b{E}_{z\sim p} [f(x)] \right) \right] + \sqrt{\frac{1}{n}}
\end{align*}

We recall the Kantorovich-Rubenstein duality theorem, which states that when the probability space is a metric space, then for any fixed $\mu > 0$,
\begin{align*}
    W_1(p, p') = \frac{1}{\mu} \sup_{\|f\|_L \leq \mu} \b{E}_{x \sim p} [f(x)] - \b{E}_{x \sim p'} [f(x)],
\end{align*}
with $\|f\|_L$ being the Lipschitz constant of $f$.

Thus, imposing a $\mu$ Lipschitz-continuity assumption on the function class gives 
\begin{align*}
    A 
    &\leq \mu \, \min_q \b{E}_{Z \sim p^n} \left[ \delta(q_Z, q) \, W_1(u_Z, u) \right]  + \sqrt{\frac{1}{n}}
\end{align*} 
which is the expected 1-Wasserstein distance between two empirical measures of $n$ samples multiplied by the Lipschitz constant $\mu$.

On the other hand, we control the difference of labels term 
by invoking once more Lemma~\ref{lemma:difference_of_labels} to deduce that $B \leq \sqrt{\frac{1}{n}}$.

Combining the bounds on $A$ and $B$ results to 
\begin{align*}
    \b{E}_Z \left[ \psi_n(Z) \right] 
    &\leq \lambda \mu \, \min_q \b{E}_{Z \sim p^n} \left[ \delta(q_Z, q) \, W_1(u_Z, u) \right] + 2 \lambda \sqrt{\frac{1}{n}} + O(\tfrac{1}{n})
\end{align*}
and thus with at least probability $1 - \delta$:
\begin{align*}
    \b{E}_{f \sim q_Z} \left[r_n(f, p)\right] 
    \leq \b{E}_{f \sim q_Z} &\left[ r_n(f, p_Z)\right] + \lambda \mu \, \min_q \b{E}_{Z \sim p^n} \left[ \delta(q_Z, q) \, W_1(u_Z, u) \right] \\ &+ 2 \lambda \sqrt{\frac{1}{n}} + (2 \sqrt{2} \lambda + \beta) \sqrt{\frac{2 \log (1 / \delta)}{n}} + O(\tfrac{1}{n}) \notag.
\end{align*}
implying also that the following holds
\begin{align}
    \b{E}_{f \sim q_Z} \left[r_n(f, p)\right] 
    &\leq \b{E}_{f \sim q_Z} \left[ r_n(f, p_Z)\right] + \lambda \mu \, \min_q \b{E}_{Z \sim p^n} \left[ \delta(q_Z, q) \, W_1(u_Z, u) \right] + O\left( \beta \sqrt{\frac{\log(1/\delta)}{n}}\right) \notag 
\end{align}
with the same probability, where we have assumed that $\lambda = O(1)$.


\subsection{Bringing everything together: the general case}

We rely on Property~\ref{th:convex-comb}, to bound the $k$-risk generalization gap in terms of those for $k=1$ and $k=n$:
\begin{align*}
    & \mathbb{P}[ r_k(f;p) - r_k(f;Z) \leq \varepsilon_1 (1 - a_{k,n}) + \varepsilon_n \, a_{k,n}] 
    = 1 - \mathbb{P}[ r_1(f;p) > r_1(f;Z) + \varepsilon_1 ] - \mathbb{P}[ r_n(f;p) > r_n(f;Z) + \varepsilon_n ]. 
 \end{align*}
We have a probability $1 - \delta$ bound 
for the $r_1$-term with 
$$
\varepsilon_1 = \lambda \mu \min_q  \b{E}_{Z \sim p^n} \left[ \delta\left(q_Z, q\right) \, 
 W_1(p, p_Z) \right] + O\left( \beta \sqrt{\frac{\log(1/\delta)}{n}}\right)
$$
and similarly for the $r_n$-term: 
$$
\varepsilon_n = \lambda \mu \min_q  \b{E}_{Z \sim p^n} \left[ \delta(q_Z, q) \, W_1(u_Z, u) \right] + O\left( \beta \sqrt{\frac{\log(1/\delta)}{n}}\right).
$$
From the former we deduce that the following inequality holds
\begin{align*}
     r_k(f;p) \leq r_k(f;Z) + \lambda \mu \min_q   \, \b{E}_{Z \sim p^n} \left[ \delta(q_Z, q) \, \left( (1 - a_{k,n})
 W_1(p, p_Z) + a_{k,n}  W_1(u, u_Z) \right] \right) + O\left( \beta \sqrt{\frac{\log(1/\delta)}{n}}\right)
 \end{align*}
with probability at least $1-2\delta$.

\section{Screening}

The standard screening error rate is obtained by re-weighting the loss so that only the predicted positives have non-negative selection probability $\pi_f(x)$ and thus contribute to the risk:
\begin{align}
	s_1(f; Z) = \sum_{z \in Z} \pi_f(x) \, l(f(x), y)
	\quad \text{with} \quad \pi_f(x) \propto w_f(x) = \frac{\alpha + f(x)}{2}. 
\end{align}
with $\alpha \geq 1$.

The probability of selecting a sample is given by  
\begin{align}
    \pi_f(x) 
    &= \frac{\frac{\alpha + f(x)}{2}}{n \frac{1}{2}\left(\alpha + \frac{1}{n} \sum_{z \in Z} f(x)\right)}
    = \frac{\alpha + f(x)}{n\left(\alpha + \frac{1}{n} \sum_{z \in Z} f(x)\right)}
\end{align}
and further, by varying $\alpha$ one may bias the selection towards positives as  
\begin{align}
    \frac{\pi_f(x|f(x)=0)}{\pi_f(x|f(x)=1)} 
    = \frac{\alpha-1}{\alpha + 1} \in [0, \sfrac{1}{2})
\end{align}
In the general case, 
\begin{align*}
    s_k(f; Z) 
    &:= \sum_{S \subset_k Z} \pi(S) \, l\left(\frac{1}{k} \sum_{z \in S} f(x), \frac{1}{k} \sum_{z \in S} y \right),
\end{align*}
where
$$
\pi_f(S) = \frac{w_f(S)}{\sum_{S \subset_k Z} w_f(S)} \propto  w_f(S) = \frac{1}{2}\left(\alpha + \frac{1}{k} \sum_{z \in S} f(x)\right).
$$
It will also be useful to notice that the denominator of $\pi_f(S)$ simplifies since
$$
\hat{\b{E}}_{S \subset_k Z} [w_f(S)] 
    = \frac{1}{\binom{n}{k}} \sum_{S \subset_k Z} \frac{1}{2}\left(\alpha + \frac{1}{k} \sum_{z \in S} f(x)\right) 
    = \frac{1}{2}\left( \alpha + \frac{1}{n} \sum_{z \in Z} f(x)\right) 
    = w_f(Z),
$$ 
with $w_f(Z)$ for $\alpha=1$ being the fraction of examples in $Z$ predicted to be positive. 

The above relation also implies:
$$
\frac{\sum_{S \subset_k Z} w_f(S)}{\binom{n}{k}} = w_f(Z) = \frac{1}{2}\left(\alpha + \frac{1}{n} \sum_{z \in Z} f(x)\right) \in [(\alpha-1)/2,1+(\alpha-1)/2].
$$ 

\subsection{Dissecting the selection mechanism}
\label{app:selectivity}

Assuming that the classifier gives confident outputs in $\{-1, 1\}$, the probability of selecting a predicted positive is given by the expected number predicted positives within a batch: 
\begin{align*}
    \pi_{f,k}(Z)
    &:= \frac{1}{2 {n \choose k}} \sum_{S \subset_k Z} \frac{w_f(S)}{w_f(Z)} \left(1 + \frac{\sum_{x\in S} f(x)}{k}\right) \\
    &\hspace{-10mm}= \frac{1}{2 {n \choose k}} \sum_{S \subset_k Z} \frac{\frac{1}{2}\left(\alpha + \frac{1}{k} \sum_{z \in S} f(x)\right)}{\frac{1}{2}\left(\alpha + \frac{1}{n} \sum_{z \in Z} f(x)\right)} \left(1 + \frac{\sum_{x\in S} f(x)}{k}\right) \\
    &\hspace{-10mm}= \frac{1}{2 {n \choose k} \left(\alpha + \frac{1}{n} \sum_{z \in Z} f(x)\right)} \sum_{S \subset_k Z} \left(\alpha + \frac{1}{k} \sum_{z \in S} f(x)\right) \left(1 + \frac{\sum_{x\in S} f(x)}{k}\right)  \\
    &\hspace{-10mm}= \frac{1}{2 \left(\alpha + \frac{1}{n} \sum_{z \in Z} f(x)\right)} \left( 
    \alpha +  (1+\alpha) \frac{1}{k {n \choose k} }\sum_{S \subset_k Z} \sum_{x\in S} f(x) + \frac{1}{k^2 {n \choose k}} \sum_{S \subset_k Z} \sum_{x\in S} \sum_{x'\in S} f(x) f(x') \right) \\
    &\hspace{-10mm}= \frac{1}{2 (\alpha + f_{\text{avg}}(Z))} \left( 
    \alpha +  (1+\alpha) f_{\text{avg}}(Z) + \frac{1}{k} \frac{1}{n}\sum_{z \in Z} f(x)^2  + \frac{k^2 - k}{k^2} \hat{\mathbb{E}}_{z \neq z'} f(x) f(x') \right) \\
    &\hspace{-10mm}= \frac{1}{2 (\alpha + f_{\text{avg}}(Z))} \left( 
    \alpha +  (1+\alpha) f_{\text{avg}}(Z) + \frac{1}{k} f_{\text{sq}}(Z) + \frac{k^2 - k}{k^2} \left( \frac{n^2}{n(n-1)} \frac{1}{n^2} \sum_{z, z' \in Z} f(x) f(x') - \frac{n}{n(n-1)} \frac{1}{n} \sum_{z \in Z} 1 \right) \right) \\
    &\hspace{-10mm}= \frac{1}{2 (\alpha + f_{\text{avg}}(Z))} \left( 
    \alpha +  (1+\alpha) f_{\text{avg}}(Z) + \frac{1}{k}f_{\text{sq}}(Z)  + \frac{1 - 1/k}{1 - 1/n} \left( f_{\text{avg}}(Z)^2 - 1/n \right) \right),
\end{align*}
where $f_{\text{avg}}(Z) = \frac{1}{n} \sum_{z \in Z} f(x)$ and $f_{\text{sq}}(Z) = \frac{1}{n} \sum_{z \in Z} f(x)^2$.

Note that $\pi_{f,1}(Z) = \frac{(1+\alpha) (f_{\text{avg}}(Z) + 1) + f_{\text{sq}}(Z)-1}{2 (\alpha + f_{\text{avg}}(Z))}$ which, assuming $f(x) \in \{-1,1\}$ for all $z \in Z$ such that $f_{\text{sq}}(Z)=1$, is 
1 for $\alpha=1$ and approaches the fraction of predicted positives $(1 + f_{\text{avg}}(Z))/2$ as $\alpha \to \infty$.
%

For a better comparison between the standard and batched cases, we fix $\alpha$ in the standard case so that the probability of selecting a predicted positive is the same as in the batched case:
\begin{align}
    \pi_{f,k}(Z) = \pi_{f,1}(Z) 
    &\Leftrightarrow \frac{\alpha_k + (1+\alpha_k) f_{\text{avg}}(Z) + \frac{1}{k}  + \frac{1 - 1/k}{1 - 1/n} \left( f_{\text{avg}}(Z)^2 - 1/n \right)}{2 (\alpha_k + f_{\text{avg}}(Z))} = \frac{(1+\alpha_1) (f_{\text{avg}}(Z) + 1)}{2 (\alpha_1 + f_{\text{avg}}(Z))} \notag \\
    &\Leftrightarrow \frac{1 + 2 f_{\text{avg}}(Z) + \frac{1}{k}  + (1 - 1/k) f_{\text{avg}}(Z)^2 }{1 + f_{\text{avg}}(Z)} = \frac{(1+\alpha_1) (f_{\text{avg}}(Z) + 1)}{\alpha_1 + f_{\text{avg}}(Z)} \tag{$n \to \infty$} \\
    &\Leftrightarrow 1 + \frac{1}{k} = \frac{1+\alpha_1}{\alpha_1} \tag{$f_{\text{avg}}(Z) = 0$} \\
    &\Leftrightarrow \alpha_1 \left(1 + \frac{1}{k}-1\right) = 1 \notag \\
    &\Leftrightarrow \alpha_1 = k \notag 
\end{align}

\subsection{Generalization bounds}

We prove the following general theorem:
\begin{theorem}
Let $l: \b{X} \times [-1, 1] \to [0,1]$ be a $\lambda$-Lipschitz continuous loss fulfilling \eqref{def:loss} and \eqref{eq:approx_jensen}.  Further, consider a $\beta$-stable learner that uses the training data $Z$ to select a distribution $q_Z$ over hypotheses $f \in \c{F}$ with Lipschitz constant at most $\mu$. The expected screening $k$-risk is upper bounded as follows
\begin{align*}
     s_k(q_Z, p)
     &\leq \frac{(\alpha + 1) \left( r_k(q_Z; p_Z) + \varepsilon_k(\mu, \lambda, p, \beta) \right) + \frac{1}{2} \, f_{\text{dev}}(q_Z, p)}{\alpha + f_{\text{avg}}(q_Z, p) }
 \end{align*}
with high probability, where 
\begin{align*}
     \varepsilon_k(\mu, \lambda, p, q, \beta) = \min_q \lambda\mu \, \b{E}_{Z} \left[ \delta(q_Z, q) \, \left( (1 - a_{k,n})
 W_1(p, p_Z) + a_{k,n}  W_1(u, u_Z) \right) \right]  + O\left( \frac{\lambda + \beta}{\sqrt{n}}\right),
\end{align*}
$W_1(u, u_Z)$ is the 1-Wasserstein distance between the empirical $u_Z$ and true measure $u$ over the unlabeled data, whereas $W_1(p, p_Z)$ measures the 1-Wasserstein distance between the labeled data. 
\label{theorem:generalization_krisk_screened}
\end{theorem}

Note that the assumption that the loss abides to~\eqref{def:loss} is needed only for $1<k<n$.

We will study
\begin{align*}
    s_k(f; p_Z) 
    &:= \sum_{S \subset_k Z} \pi_f(S) \, l\left(\frac{1}{k}\sum_{z \in S} f(x), \frac{1}{k}\sum_{z \in S} y \right) \quad \text{with} \quad \pi_f(S) = \frac{\alpha + \frac{1}{k}\sum_{z \in S} f(x)}{2 \, {n\choose k} w_f(Z)}
\end{align*}
Set $g(f; p_Z) := s_k(f; p_Z) w_f(Z)$ and $w := \b{E}_{Z'\sim p^n,f\sim q_Z}[w_f(Z')]$.

We continue to upper bound the expected screening risk w.r.t. the expectation of $g(f; Z)$:
\begin{align*}
    s_k(q_Z; p)
    &= \frac{1}{w} \sum_{f \in \c{F}} q_Z(f) g(f; p) + \sum_{f \in \c{F}} q_Z(f) \sum_{Z'} p(Z') g(f,Z) \left( \frac{1}{w_f(Z')} - \frac{1}{w} \right) \\  
    &\hspace{-20mm}= \frac{1}{w} \sum_{f \in \c{F}} q_Z(f) g(f; p) + \sum_{f \in \c{F}} q_Z(f) \sum_{Z'} p(Z') \left(\sum_{S \subset_k Z'} \frac{\alpha + \frac{1}{k}\sum_{z' \in S'} f(x')}{2 \, {n\choose k}} \, l\left(\frac{1}{k}\sum_{z' \in S'} f(x'), \frac{1}{k}\sum_{z' \in S'} y' \right) \right) \left( \frac{1}{w_f(Z')} - \frac{1}{w} \right) \\
    &\hspace{-20mm}\leq \frac{1}{w} \sum_{f \in \c{F}} q_Z(f) g(f; p) + \sum_{f \in \c{F}} q_Z(f) \sum_{Z'} p(Z') \left(\sum_{S \subset_k Z'} \frac{\alpha + \frac{1}{k}\sum_{z' \in S'} f(x')}{2 \, {n\choose k}} \right) \left| \frac{1}{w_f(Z')} - \frac{1}{w} \right|_+ \\
    &\hspace{-20mm}= \frac{1}{w} \left( \sum_{f \in \c{F}} q_Z(f) g(f; p) + \sum_{f \in \c{F}} q_Z(f) \sum_{Z'} p(Z') \left| w - w_f(Z') \right|_+ \right) \\
    &\hspace{-20mm}= \frac{1}{w} \left( \sum_{f \in \c{F}} q_Z(f) g(f; p) + \frac{1}{2}\b{E}_{Z'\sim p^n,f\sim q_Z}[|w - w_f(Z')|] \right),
\end{align*}
where the last step follows from that $\b{E}_{}[w - w_f(Z')] = 0$ and, moreover, that for any zero mean random variable $x$ we have $\b{E}[\left|x\right|_+] = \frac{\b{E}[\left|x\right|]}{2}$ since $\b{E}[\left|x\right|_+] = \b{E}[\left|x\right|_-]$.
%
%
We also note that 
$$
w = \b{E}_{Z'\sim p^n,f\sim q_Z}[w_f(Z')] = \frac{1}{2} \left(\alpha + f_{\text{avg}}(q_Z, p) \right) 
$$
and further
\begin{align*}
    \frac{1}{2}\b{E}_{Z'\sim p^n,f\sim q_Z}[|w - w_f(Z')|]
    &= \frac{1}{2}\b{E}_{Z'\sim p^n,f\sim q_Z}[|\b{E}_{Z''\sim p^n,f\sim q_Z}[w_f(Z'')]- w_f(Z')|]\\
    &= \frac{1}{2}\b{E}_{Z'\sim p^n,f\sim q_Z}[|\b{E}_{Z''\sim p^n,f\sim q_Z}[\frac{1}{2}\left(\alpha + \frac{1}{n} \sum_{z \in Z''} f(x)\right)]- \frac{1}{2}\left(\alpha + \frac{1}{n} \sum_{z \in Z'} f(x)\right)|] \\
    &= \frac{1}{4} \, \b{E}_{Z'\sim p^n,f\sim q_Z}[|\b{E}_{Z''\sim p^n,f\sim q_Z}[\frac{1}{n} \sum_{z \in Z''} f(x)]- \frac{1}{n} \sum_{z \in Z'} f(x)|] \\
    &= \frac{1}{4} \, \b{E}_{Z'\sim p^n, f\sim q_Z} \left[\left| \b{E}_{p_{Z'}}[f(x)] - f_{\text{avg}}(q_Z, p) \right|\right] \\
    &= \frac{1}{4} \, f_{\text{dev}}(q_Z, p)
\end{align*}
with
\begin{align*}
    f_{\text{avg}}(q_Z, p) = \b{E}_{z\sim p,f \sim q_Z}[f(x)] 
    \quad \text{and} \quad
    f_{\text{dev}}(q_Z, p) = \b{E}_{Z'\sim p^n, f\sim q_Z} \left[\left| \b{E}_{p_{Z'}}[f(x)] - f_{\text{avg}}(q_Z, p) \right|\right]
\end{align*}

The re-normalized screening risk is upper bounded by the $k$-risk:
\begin{align*}
    g(f; p_Z) 
    &= \sum_{S \subset_k Z} \frac{\alpha + \frac{1}{k}\sum_{z \in S} f(x)}{2 \, {n\choose k}} \, l\left(\frac{1}{k}\sum_{z \in S} f(x), \frac{1}{k}\sum_{z \in S} y \right) \\
    &\leq \sup_S \{ \frac{\alpha + \frac{1}{k}\sum_{z \in S} f(x)}{2} \} \, \frac{1}{ {n\choose k}} \sum_{S \subset_k Z}  l\left(\frac{1}{k}\sum_{z \in S} f(x), \frac{1}{k}\sum_{z \in S} y \right) \\
    &= \frac{\alpha + 1}{2} \, r_k(f; p_Z) 
\end{align*}
and similarly $g(q_Z; p) \leq \frac{\alpha + 1}{2} \, r_k(q_Z; p)$.
Thus, by applying the result of Theorem~\ref{theorem:generalization_krisk}, one may deduce that 
\begin{align*}
    g(q_Z, p)
    &\leq \frac{\alpha + 1}{2} \, r_k(q_Z; p)
    \leq \frac{\alpha + 1}{2} \left( r_k(q_Z; p_Z) + \varepsilon_k(\mu, \lambda, p, \beta) \right) 
\end{align*}
with 
\begin{align*}
     \varepsilon_k(\mu, \lambda, p, q, \beta) = \min_q \lambda\mu \, \b{E}_{Z} \left[ \delta(q_Z, q) \, \left( (1 - a_{k,n})
 W_1(p, p_Z) + a_{k,n}  W_1(u, u_Z) \right) \right]  + O\left( \frac{\lambda + \beta}{\sqrt{n}}\right)
\end{align*}
Putting everything together results in the following generalization bound
\begin{align}
    s_k(q_Z; p)
    &\leq \frac{1}{w} \left( g(q_Z; p) + \frac{1}{2}\b{D}_{Z'\sim p^n,f\sim q_Z}[w_f(Z')] \right)  \notag \\
    &\leq \frac{\frac{\alpha + 1}{2} \left( r_k(q_Z; p_Z) + \varepsilon_k(\mu, \lambda, p, \beta) \right) + \frac{1}{4} \, f_{\text{dev}}(q_Z, p)}{\frac{1}{2} \left(\alpha + f_{\text{avg}}(q_Z, p) \right) } \notag \\
    &= \frac{(\alpha + 1) \left( r_k(q_Z; p_Z) + \varepsilon_k(\mu, \lambda, p, \beta) \right) + \frac{1}{2} \, f_{\text{dev}}(q_Z, p)}{\alpha + f_{\text{avg}}(q_Z, p) },
\end{align}
that holds with high probability.

In the realizable case, the bound simplifies to
\begin{align}
    s_k(q_Z; p)
    &\leq \frac{ (\alpha + 1) \, \varepsilon_k(\mu, \lambda, p, q, \beta) + \frac{1}{2} \, f_{\text{dev}}(q_Z, p)}{\alpha + f_{\text{avg}}(q_Z, p) },
\end{align}
with 
$$
   \varepsilon_k(\mu, \lambda, p, q, \beta) = \min_q \lambda\mu \, \b{E}_{Z} \left[ \delta(q_Z, q) \, \left( (1 - a_{k,n})
 W_1(p, p_Z) + a_{k,n}  W_1(u, u_Z) \right) \right]  + O\left( \frac{\lambda + \beta}{\sqrt{n}}\right)
$$
whereas for $k=1$ we further have
$$
\varepsilon_1(\mu, \lambda, p, q, \beta) = \min_q \lambda\mu \, \b{E}_{Z} \left[ \delta(q_Z, q) \, 
 W_1(p, p_Z) \right]  + O\left( \frac{\lambda + \beta}{\sqrt{n}}\right)
$$

When $f_{\text{avg}}(q_Z, p) = o(1)$ and $f_{\text{dev}}(q_Z, p) = o(1)$

\subsection{Understanding the effect of batching under the same selectivity}

Let us contrast the obtained bound to the generalization error for $k=1$ and $\alpha = k$ which gives the same probability for selecting predicted positives: 
\begin{align}
    s_1(q_Z; p)
    &\leq \frac{ (1 + 1/k) \, \varepsilon_1(\mu, \lambda, p, q, \beta) + \frac{1}{2 k} \, f_{\text{dev}}(q_Z, p)}{1 + f_{\text{avg}}(q_Z, p)/k }
    = \varepsilon_1(\mu, \lambda, p, q, \beta) + O\left( \frac{1}{k} \right)
\end{align}
whereas for general $k$ and $\alpha=1$ we have
\begin{align}
    s_k(q_Z; p)
    &\leq \frac{ 2\, \varepsilon_k(\mu, \lambda, p, q, \beta) + \frac{1}{2} \, f_{\text{dev}}(q_Z, p)}{1 + f_{\text{avg}}(q_Z, p) }
\end{align}
with $\varepsilon_k(\mu, \lambda, p, q, \beta)$ as above. 
%

\subsection{Relation between the convergence of labeled and unlabeled empirical measures}

\begin{property}
The labeled empirical measure converges at most as fast as the unlabeled empirical measure: $W_1(p, p_Z) \geq W_1(u, u_Z)$.
Moreover, for any unlabeled measures $u$ and $u_Z$ and any labeling function $y = f^*(x)$ of Lipschitz constant $\lambda_*$, we have $W_1(p, p_Z) \geq \lambda_* \,  W_1(u, u_Z)$.  
\label{property:convergence_labeled_unlabeled}
\end{property}

\begin{proof}
For the general result, we consider the distance between labeled measures as given by the Kantorovich-Rubenstein duality theorem:
\begin{align*}
    W_1(p, p_Z) 
    &=\frac{1}{2} \sup_{\|h\|_L \leq 2} \left( \b{E}_{z \sim p} [h(z)] - \b{E}_{z \sim p_Z} [h(z)] \right) 
\end{align*}
and proceed to restrict the set of functions in the supremum to those that can be written as $h(x,y) = g(x)$ with $\|g\|_L \leq 1$:
\begin{align*}
    W_1(p, p_Z) 
    &\geq \sup_{g : \|g\|_L \leq 1} \left( \b{E}_{z \sim p} [g(x)] - \b{E}_{x \sim p_Z} [g(x)] \right) \\
    &= W_1(u, u_Z),
\end{align*}
To understand how much larger the labeled and unlabeled distances can be, denote by $f^*$ the function of Lipschitz constant $\lambda$ for which the Kantorovich-Rubenstein duality theorem yields:
\begin{align*}
    W_1(u, u_Z) = \frac{1}{\lambda} \sup_{\|f\|_L \leq \lambda} \left( \b{E}_{x \sim u} [f(x)] - \b{E}_{x \sim u_Z} [f(x)] \right) 
    = \frac{1}{\lambda} \left( \b{E}_{x \sim u} [f^*(x)] - \b{E}_{x \sim u_Z} [f^*(x)] \right)  
\end{align*}
We will consider labels $y = f^*(x)$ and restrict to functions $h(z) = g(y)$ to obtain
\begin{align*}
    W_1(p, p_Z) 
    &= \sup_{\|h\|_L \leq 1} \b{E}_{z \sim p} [h(z)] - \b{E}_{z \sim p_Z} [h(z)]  \\
    &\geq \sup_{\|g\|_L \leq 1} \left( \b{E}_{y \sim v} [g(y)] - \b{E}_{y \sim v_Z} [g(y)] \right) \tag{$v,v_Z$ are probability measures of labels} \\
    &= \sup_{g : \|g\|_L \leq 1} \left( \b{E}_{x \sim p} [g(f^*(x))] - \b{E}_{x \sim p_Z} [g(f^*(x))] \right)  \tag{since $y = f^*(x)$} \\
    &\geq \b{E}_{x \sim p} [f^*(x)] - \b{E}_{x \sim p_Z} [f^*(x)]   \tag{restrict to $g(y) = y$} \\
    &= \lambda \, W_1(u, u_Z)
\end{align*}
as claimed.
\end{proof}

\end{document}